\RequirePackage{fix-cm}

\documentclass[twocolumn]{svjour3}          
\smartqed  
\usepackage{url}
\usepackage{natbib}
\usepackage{graphicx}
\usepackage{times}
\usepackage{epsfig}
\usepackage{graphicx}
\usepackage{tabularx}
\usepackage{amsmath}
\usepackage{amssymb}
\usepackage{comment}
\usepackage{makecell}
\usepackage{pifont}
\usepackage{multirow}
\usepackage{subfigure}
\usepackage[misc]{ifsym}
\usepackage{afterpage}
\usepackage{setspace}
\usepackage[breaklinks=true]{hyperref}
\usepackage{breakcites}
\usepackage{seqsplit}

\usepackage{wrapfig}
\usepackage{hyphenat}
\usepackage{soul,color}
\usepackage{amsopn}
\usepackage{amsmath}
\usepackage{amssymb}
\usepackage{tabularx}
\usepackage{subfigure}

\definecolor{Red}{cmyk}{0,1,1,0}
\definecolor{Green}{cmyk}{1,0,1,0}
\definecolor{Cyan}{cmyk}{1,0,0,0}
\definecolor{Purple}{cmyk}{0.45,0.86,0,0}
\definecolor{Rosolic}{cmyk}{0.00,1.00,0.50,0}
\definecolor{Blue}{cmyk}{1.00,1.00,0.00,0}
\definecolor{BlueViolet}{cmyk}{0.86,0.91,0,0.04}
\definecolor{NavyBlue}{cmyk}{0.94,0.54,0,0}

\newcommand{\chenNew}[1]{{#1}}
\newcommand{\pang}[1]{{#1}}

\newcommand{\myparagraph}[1]{\vspace{0.1em}\noindent\textbf{#1}}
\journalname{arXiv 2021.07.15 https://chenxin.tech/SportsCap.html}
\begin{document}

\title{SportsCap: Monocular 3D Human Motion Capture and Fine-grained Understanding in Challenging Sports Videos 
}

\author{Xin Chen$^{1,2,3}$      \and
        Anqi Pang$^{1,2,3}$ \and
        Wei Yang$^4$  \and
        Yuexin Ma$^1$ \and
        Lan Xu$^{1 *}$    \and
        Jingyi Yu$^{1 *}$
}



\institute{Xin Chen \at
              \email{chenxin2@shanghaitech.edu.cn}
           \and
           Anqi Pang \at
              \email{pangaq@shanghaitech.edu.cn}
           \and
           Wei Yang \at
              \email{wyangcs@udel.edu}
           \and
           Yuexin Ma \at
              \email{mayuexin@shanghaitech.edu.cn}
           \and
           Lan Xu \at
              \email{xulan1@shanghaitech.edu.cn}
           \and
           Jingyi Yu \at
              \email{yujingyi@shanghaitech.edu.cn}
           \\
           $^1$ Shanghai Engineering Research Center of Intelligent Vision and Imaging
           School of Information Science and Technology ShanghaiTech University, Shanghai, China
           \\
           $^2$ Shanghai Institute of Microsystem and Information Technology, Chinese Academy of Sciences, Shanghai, China
           \\
           $^3$ University of Chinese Academy of Sciences, Beijing, China
           \\
           $^4$ School of Computer Science \& Technology, Huazhong University of Science and Technology, Wuhan, China
           \\
           $^*$ The corresponding authors are Lan Xu and Jingyi Yu.
}
\date{Received: date / Accepted: date}
\maketitle
\begin{abstract}
Markerless motion capture and understanding of professional non-daily human movements is an important yet unsolved task, which suffers from complex motion patterns and severe self-occlusion, especially for the monocular setting.
In this paper, we propose SportsCap -- the first approach for simultaneously capturing 3D human motions and understanding fine-grained actions from monocular challenging sports video input.
Our approach utilizes the semantic and temporally structured sub-motion prior in the embedding space for motion capture and understanding in a data-driven multi-task manner.
To enable robust capture under complex motion patterns, we propose an effective motion embedding module to 
recover both the implicit motion embedding and explicit 3D motion details via a corresponding mapping function as well as a sub-motion classifier.
Based on such hybrid motion information, we introduce a multi-stream spatial-temporal Graph Convolutional Network(ST-GCN) to predict the fine-grained semantic action attributes, and adopt a semantic attribute mapping block to assemble various correlated action attributes into a high-level action label for the overall detailed understanding of the whole sequence, so as to enable various applications like action assessment or motion scoring.
Comprehensive experiments on both public and our proposed datasets show that with a challenging monocular sports video input, our novel approach not only significantly improves the accuracy of 3D human motion capture, but also recovers accurate fine-grained semantic action attribute.


\end{abstract}

\section{Introduction}

The past ten years have witnessed a rapid development of markerless motion capture and understanding for human daily activities, which benefits various real-world applications such as immersive VR/AR experience, action quality assessment \citep{Pan_2019_ICCV} and vision-based robotics \citep{ran2017convolutional}.
How to further capture professional non-daily human motions and provide fine-grained analysis has recently received substantive attention.

%

\begin{figure*}[t]
  \centering
  \includegraphics[width=0.95\linewidth]{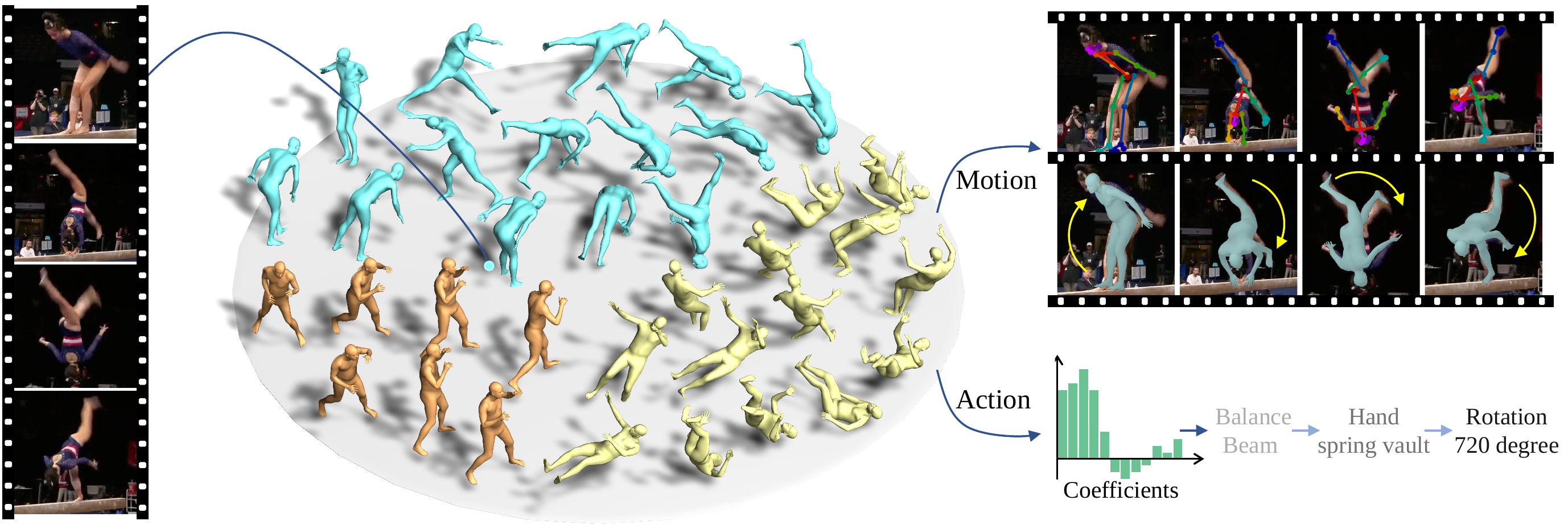}
  \caption{SportsCap: A multi-task approach for 3D motion capture and action understanding of challenging sports videos. We collect a sport-related motion capture dataset to build the Sports Pose Embedding Spaces on specific sports, like balance beam (Blue), boxing (Orange), high jump (Yellow), and other sports. This Sports Pose Embedding Spaces achieve significant superiority on challenging motion capture and encode semantic meanings (see Fig. \ref{fig:Coeff_Vis}) for action parsing tasks.
  }
  \label{fig:teaser}
\end{figure*}

In this paper, we focus on markerless motion capture and fine-grained understanding for challenging professional human movements which are essential for many applications such as training and evaluation for gymnastics, sports, and dancing.
However, these professional movements like diving and balance-beam suffer from complex motion patterns and severe self-occlusion, especially under the monocular setting, leading to inferior results and impractical usage of existing 3D motion capture~\citep{xiao2018simple,kocabas2020vibe} and 2D pose detection approaches~\citep{cao2018openpose}.
When motion capture is unreliable, further motion analysis is even more challenging, which aims to provide both  
mid-level sub-motion categories and detailed semantic descriptions for each sub-motion or the whole motion sequence at the finest granularity.
On the other hand, even though it's natural to split such challenging sports movements into sub-motions due to their repeatability and self-similarity, the literatures on utilizing such sub-motion prior to strengthening the motion capture and understanding are sparse.
Moreover, most existing action understanding solutions~\citep{Parmar_2019_CVPR,shao2020finegym} are limited to the pure high-level action assessment, where the abundant 3D motion capture information of sub-motions has been ignored. 

To tackle these challenges, we propose \textit{SportsCap} -- the first joint 3D motion capture and fine-grained understanding approach for various challenging sports movements from only a single RGB video input (see Fig.~\ref{fig:teaser} for an overview).
With the aid of mid-level sub-motion embedding analysis of plausible motion manifold, SportsCap explores and validates the mutual gain between 3D motion capture and fine-grained motion understanding.
Our novel pipeline not only achieves significant superiority to previous capture methods for challenging motions, but also provides accurate fine-grained semantic assessment simultaneously for motion understanding, whilst still maintaining a monocular setup. 

More specifically, we formulate this joint human motion capture and understanding problem in a multi-task learning framework.
To this end, we first introduce a motion embedding space to model the manifold of plausible human poses for each sub-motion via the principal component analysis (PCA) technique.
Then, for the motion capture task, an effective motion embedding network is proposed to estimate the per-frame implicit embedding parameters so as to recover the 3D motion details via a corresponding mapping function as well as a sub-motion classifier. 
Our motion capture scheme leverages the rich semantic and temporally structural prior of sub-motions in the motion embedding space to tackle the severe occlusion and depth ambiguities inherent to the monocular sports video input.
For further motion understanding task, we predict the fine-grained semantic action attributes using a spatial-temporal 
Graph Convolutional Network(ST-GCN), based on both the original motion embedding stream and the recovered 3D detailed motion stream of the whole video clip.
Our novel multi-stream ST-GCN module encodes both the implicit and explicit motion information from the previous capture stage for more accurate action attribute parsing.
Finally, a semantic attribute mapping block is adopted to assemble various correlated action attributes into a  
high-level action label for the whole sequence (i.e. the diving number for the diving motion), which provides an extra overall detailed understanding of the whole video to enable various applications like action assessment or motion scoring. 
To summarize, the main contributions of SportsCap include:
\begin{itemize} 
  \setlength\itemsep{0em}
  \item We propose a novel join human 3D motion capture and motion understanding scheme in a data-driven multi-task manner under the monocular setting, achieving significant superiority to existing state-of-the-arts.
  
  \item By utilizing the semantic and temporally structured motion prior in the embedding space, we propose a novel motion embedding module, as well as an effective multi-stream ST-GCN module to reconstruct both detailed 3D motions and accurate fine-grained actions, attributes simultaneously. 
  
  \item We make available our Sports Motion and Recognition Tasks (SMART) dataset, 
  consisting of various challenging sports video clips with manually annotated poses and fine-grain action labels as well as the relevant ground truth 3D poses for motion embedding analysis.
\end{itemize}

\section{Related Work}



\myparagraph{{Pose and Shape Estimation}}
aims to recover the underlying kinematic structure of a person. The results of these work can be 2D/3D poses or 3D human models that match the image/video observations. Earlier methods adopted geometric constraints~\citep{yang2011articulated} to construct poses.
Recently, with the success of deep neural networks in many computer vision tasks, many deep learning-based pose estimation approaches \citep{cao2018openpose, pishchulin2016deepcut, raaj2019efficient, sun2019deep,sun2018integral, tang2018quantized} have achieved remarkable performance. OpenPose~\citep{cao2018openpose} employs Part Affinity Fields (PAF) to support bottom-up estimation. \citet{sun2019deep} exploits multi-scale high-resolution networks to improve feature representation. \citet{li2019pose2body} jointly optimize human pose and segmentation. However, such methods focus on regular movements and actions and have limitations to handling professional sports, which consist of more complex poses and occlusions in monocular videos. A few recent approaches aim to tackle special actions. \citet{luvizon20182d} propose a semantic-based, multi-task learning framework, and \citet{bertasius2018learning} tailors a predictor specific to certain actions.
ChallenCap \citep{Challencap2021} captures challenging 3D human motions with the multi-modal references. These approaches do not consider rich semantic information embedded in sports, and the structure constraints within sub-motions. In contrast, our approach explicitly uses the underlying semantic and ordering rules in sport to reduce the complexity of the problem. And we utilize PCA to capture the similarities of poses in each sub-motion and constrain estimated poses in reasonable forms to further improve the accuracy.

Traditional 3D human estimation methods either use multi-camera dome systems \citep{kanade1997virtualized, collet2015high, Suo_2021_CVPR} or exploit the RGB-D sensors \citep{dou2016fusion4d, newcombe2015dynamicfusion,xu2019unstructuredfusion}, and recover the human geometry via multi-view stereo and point cloud fusion. With the advance in parametric 3D human body models and deep neural networks, especially with the emergence of the SCAPE \citep{anguelov2005scape}, SMPL \citep{SMPL:2015}, SMPL-X \citep{SMPL-X:2019}, recovering the human shape from a single viewpoint image/video becomes more and more popular. SMPLify-X \citep{SMPL-X:2019} fits the face, hand, and body parts of the SMPL-X model to images with pre-estimated 2D poses. TightCap \citep{chen2019tightcap} captures both the body shape and dressed garments with a single 3D human scan. HMR \citep{HMR} proposes an end-to-end framework to regress the pose and shape parameters of human model directly from a single image supervised by an adversarial prior. Similarly, the VIBE \citep{kocabas2020vibe} leverages the human pose data from a motion capture dataset, AMASS \citep{mahmood2019amass}, and develops an adversarial framework to discriminate between real human motions and the produced temporal pose and shape. Recovering the human shape from competitive sports images/videos is even more challenging. The aforementioned methods rely on 2D poses to regress the human pose and shape parameters, while athletes in competitive sports exhibit highly complex poses and fast motions that won't appear in daily activities. We tackle the problem through embedding these highly complex but standard human poses (typical poses in many sports guidelines, like twisting in diving, turning in balance beam) to parametric space. 

\myparagraph{{Action Parsing}} can be categorized into short vs. long dynamics, depending on the length of the motion patterns. For short term dynamics, \citet{karpathy2014large} uses 2D CNNs to learn deep appearance features and conduct frame-level classification. IDT~\citep{feichtenhofer2016convolutional} extends the technique with shallow motion features and~\citet{hussein2019timeception} uses 3D CNNs such as C3D~\citep{tran2015learning} to capture spatial-temporal patterns of consecutive frames within the sequence. For long term dynamics, TRN~\citep{zhou2018temporal} exploits temporal dependencies across video frames over multiple hierarchies. TRN~\citep{zhou2018temporal} proposes a multi-stream architecture to extract even richer temporal features. LTC~\citep{varol2017long} treats the temporal resolutions as a substitute to temporal windows whereas \citet{hussein2019timeception} conducts long-range action recognition. We observe that competitive sports, like diving and gym, are always a mixture of long and short dynamics: actions such as twisting or somersaults map to short dynamics whereas the complete dive, with a corresponding dive number, map to long dynamics. We hence combine the advantages of short and long dynamics techniques.

To specifically tackle sports videos,~\citet{kanojia2019attentive} proposes an attentive guided LSTM-based neural network for fine-grained motion recognition. \citet{pishchulin2014fine} combines the dense motion trajectories and pose estimations to improve recognition accuracy.~\citet{choutas2018potion} represents the movement of semantic keypoints as a color encoded trajectory map, called PoTion, and subsequently conducts classification on the PoTion. In a similar vein,~\citet{fani2017hockey} stacks the poses features generated by an hourglass network into a reference frame and then performs the fine-grained action recognition from hockey sports videos.~\citet{Pan_2019_ICCV} builds a joint relation graph to model both the joint relations within a time step and across two immediate time steps. Nevertheless, though these approaches rely on the joint motions for action recognition, they ignore the patterns of the human body motion in certain activities. In contrast, we observe that the joint motion within a fine-grained action tends to be regular in competitive sports. Hence, we adopt a fine-grained manner to model the pose in each fine-grained action and finally resorts to the recent Graph Convolutional Network (GCN)~\citep{defferrard2016convolutional, henaff2015deep, li2018spatio, li2018adaptive,shi2019two} for spatial-temporal representations.

\chenNew{\myparagraph{{Sports}} images/videos provide more challenging motions and environment for learning tasks. These tasks are numerous, ranging from correcting athletes' movements \cite{Pan_2019_ICCV,shao2020finegym} for improving their performance to digitally producing 3D avatars \citep{rematas2018soccer,zhu2020reconstructing} for video games and feature films. \cite{rematas2018soccer} built a CNN-based system to transform a soccer video into a moving 3D reconstruction, while \cite{zhu2020reconstructing} reconstruct skinned models of basketball players with a single input photo of a clothed player. For accurate tracking during big sports, such as soccer and basketball, \cite{chen2019sports,sha2020end} propose automatic approaches of camera calibration with semantic segmentation and detected edge of sports environment, like field marking. \cite{bertasius2018egocentric,su2017predicting,bertasius2017baller} propose the learning-based approaches to estimate motion, behaviors, and performance assessment of basketball players. Moreover, many official sports organizing committees provide the detailed rules of standard poses and assessment approaches, like Federation Internationale de Natation (FINA) for diving and Fdration Internationale de Gymnastique (FIG) for gymnastics.}

\myparagraph{{Dataset}} is the basis for deep learning-based motion estimation and action parsing methods. There are some large-scale human image/video datasets, such as COCO~\citep{lin2014microsoft} and MPII~\citep{andriluka14cvpr}. They mainly focus on motions in daily motions. Competitive sports video understanding relies heavily on available sports datasets.~\citet{Zhang_2013_ICCV} proposes a simple motion dataset of 15 actions with annotated body joints but no action labels.~\citet{parmar2019action, Parmar_2019_CVPR} presents the MTL-AQA dataset that exploits multi-task networks along with a caption generation model to simultaneously assess the move and produce a caption.~\citet{li2018resound} proposes the Diving48 dataset for competitive diving video understanding. The UCF101 dataset~\citep{soomro2012dataset} contains 101 classes of in-the-wild actions and the ActivityNet \citet{caba2015activitynet} covers a wide range of complex human activities in daily living. More recently,~\citet{shao2020finegym} proposes the FineGym dataset which contains 10 event categories, 303 competition records and provides coarse-to-fine annotations both temporally and semantically. Competitive sports videos contain both rich semantic action information and strict human body motions. Similar to the FineGym and Diving48, our SMART dataset contains per-frame annotated action labels. In addition to the fine-grained semantic labels, we further add manually annotated human pose, MoCap pose space of each fine-grained action, and action assessment from professional referees. To our knowledge, the SMART dataset is the only one that provides the fine-grained semantic labels, 2D and 3D annotated poses, and assessment information.

\begin{figure*}[t]
  \centering
  \includegraphics[width=\linewidth]{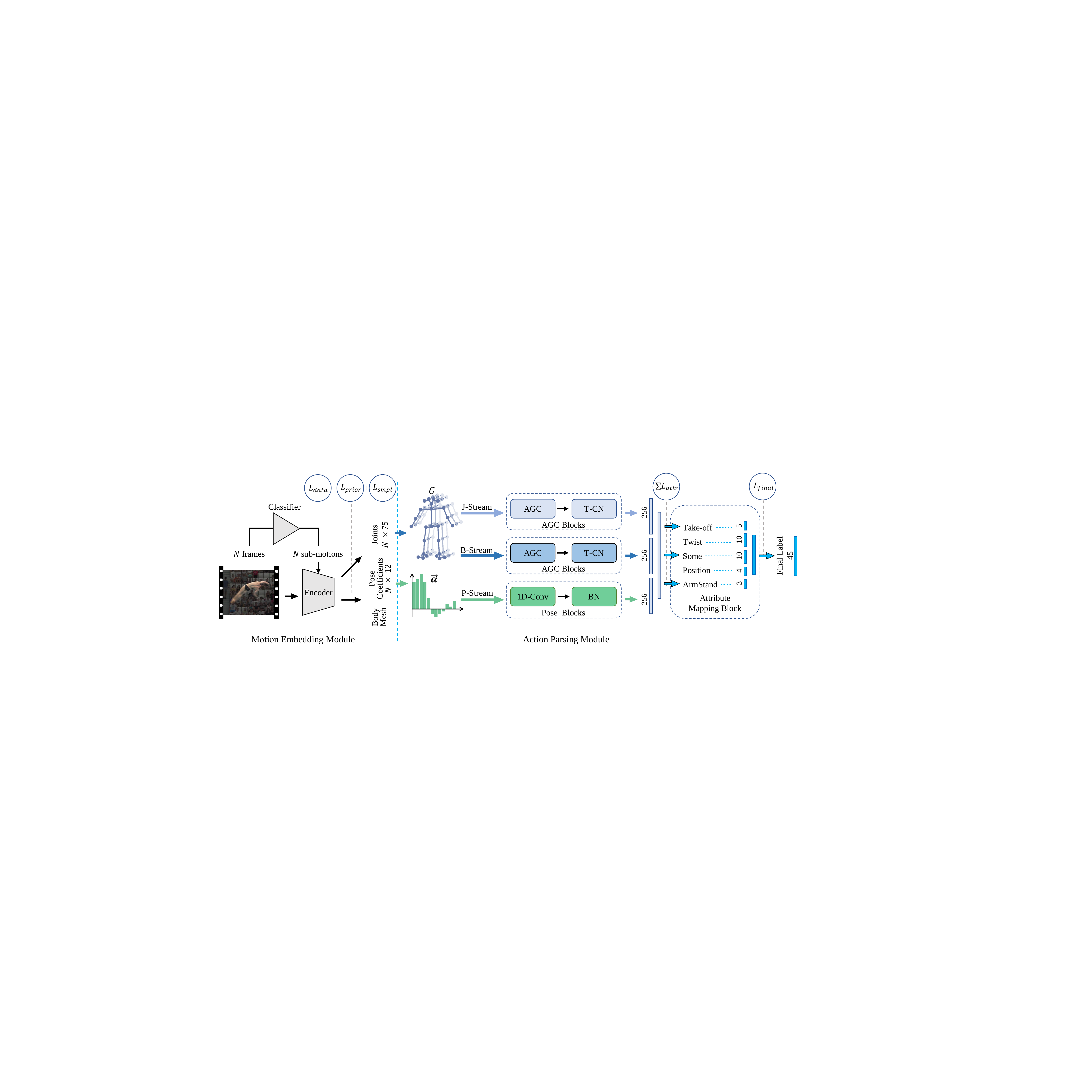}
  \caption{Our SportsCap is composed of two main components: the Motion Embedding Module and the Action Parsing Module. Motion Embedding Module estimates motion embedding information, 3D joints, and 3D body meshes, while Action Parsing Module predicts the fine-grained semantic attributes and final action labels of sports.}
  \label{fig:pipeline}
\end{figure*}

\section{Overview}
This paper aims to reconstruct both the 3D human motion and the corresponding fine-grained action attributes from monocular professional sports video input. 
To handle this challenging problem, our SportsCap splits each professional motion into a sequence of elementary sub-motions, and utilizes the motion manifold prior of these sub-motions in a multi-task learning framework, as illustrated in Fig.~\ref{fig:teaser}.
\chenNew{Our approach not only captures the fine 3D motion details for each sub-motion, but also provides detailed motion understanding attributes, such as the action type and rotation angle in Fig. \ref{fig:teaser}.}
To model this motion capture and understanding problem in a data-driven manner, we collect a new Sports Motion and Recognition Tasks (SMART) dataset. It contains various challenging sports video clips, manually annotated ground truth poses and fine-grain action labels, and the corresponding relevant 3D poses captured via a motion capture system.
A brief introduction of our pipeline's two main components is provided as follows, which explores and proves the mutual gain between 3D motion capture and fine-grained motion understanding.

\myparagraph{Motion Embedding Module.} 
To handle a challenging sports video, we first propose a motion embedding space to model the manifold of plausible human poses for each sub-motion via the PCA technique. 
Based on such embedding prior, we further introduce a novel network to estimate the per-frame implicit motion embedding parameters so as to recover the 3D motion details, including the pose, shape parameters of the human statistical model SMPL~\cite{SMPL:2015} and camera parameters.  
Our embedding module consists of a sub-motion classifier, a CNN encoder to regress the embedding, and the corresponding mapping function from the embedding space to the 3D motion output (see Sec. \ref{Motion Embedding Module}).

\myparagraph{Action Parsing Module.} 
For further motion understanding tasks, we predict the fine-grained action attributes for the whole motion sequence using a novel multi-stream ST-GCN module, which makes full use of both the implicit pose embedding and the explicit 3D joints from the previous capture stage.
%
We further propose a semantic attribute mapping block to map the predicted attributes to the final action label, which enables various applications such as action number prediction (like diving or gymnastics number) for motion scoring and action assessment.

\section{Technical Details of SportsCap}
Fig.~\ref{fig:pipeline} shows the architecture of our SportsCap, which takes a sports video as input and reconstructs both 3D motion details and accurate action attributes in an end-to-end multi-task manner. 
We assume the input video clip corresponds to the complete sports motion and split it into several sub-motions (such as Fig. \ref{fig:DivePoseSpace}), which are segments that correspond to its sports stages similar to previous work~\citep{hu2019see}. 
%
Our SportsCap consists of two modules for both per-frame 3D pose/shape reconstruction and action understanding, such as fine-grained labeling and action assessment. 
The recovered pose and shape parameters can be further applied to drive a 3D parametric human model to conduct the same sports movement in 3D. 
For motion capturing, we construct their corresponding motion embedding functions for respective segments where each frame is fed to its respective encoder to obtain its motion embedding information, joints, and bones. 
For action labeling, we construct a multi-stream ST-GCN for multi-task action attribution prediction, which takes the coefficients, joints, and bones as input.
Our ST-GCN contains an attributes mapping block that assembles action attributes into the final label, which indicates an action number or score.

\begin{figure*}[t]
    \centering
  \includegraphics[width=\linewidth]{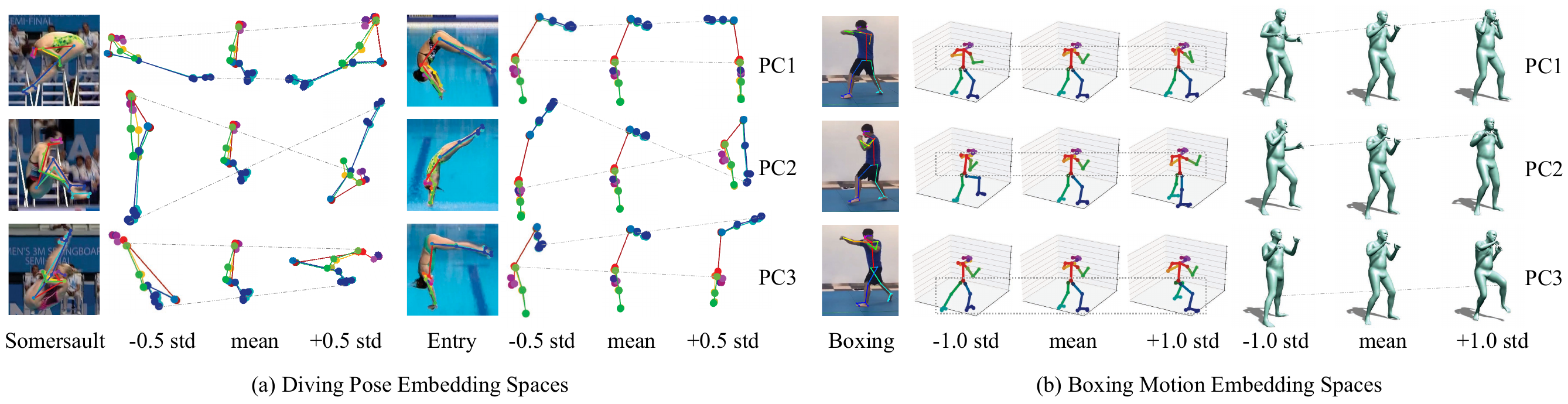}
  \caption{Visualization on motion embedding pose spaces of 2D poses, 3D poses, and pose parameters of SMPL by changing the pose coefficients: (a) 2D Pose embedding spaces for somersault and entry (two sub-motions of competitive diving). (b) Motion embedding spaces of 3D poses and pose parameters of SMPL for boxing. For each sub-motion, from left to right, we show the input frame and the first three principal components(PC) within 0.5 or 1 deviations of pose coefficients from the mean. The lines connecting the corresponding elements within the component indicate the linear change according to the basis.}
  \label{fig:feature}
\end{figure*}

\subsection{Motion Embedding and Capturing}
\label{Motion Embedding Module}
Professional poses in sports have complex structure information and always bring occlusions in monocular videos, which impose significant challenges to existing pose/shape estimators such as OpenPose \citep{cao2018openpose}, SimpleBaseline \citep{xiao2018simple}, HMR \citep{HMR} and VIBE \citep{kocabas2020vibe}. Fig.~\ref{fig:PoseResults} shows some typical results. This is partially due to the pose variants in sports. More importantly, those approaches do not explore the specific semantic and structural constraints in sports. \chenNew{We thus present a novel motion embedding space of each specific sport, to model the manifold of plausible human poses for each sub-motion via the PCA technique, and use the motion embedding network to estimate the per-frame implicit embedding parameters so as to recover the 3D motion details.}


\label{sms}We first recognize that the complete sport move in profession always follows several stages, called sub-motions. \chenNew{A sub-motion is a segmentation of the video sequence in the temporal domain, according to movement regularity and semantically meaningful.} For example, boxing action can be segmented into three sub-motions: punching, kicking, and dodging, and the diving action has four sub-motions, as shown in Fig. \ref{fig:DivePoseSpace}). In each sub-motion, the poses exhibit high resemblance across athletes, e.g., divers straighten their bodies in twisting while curling up in somersault. To achieve this, we use an accurate and effective classifier, WS-DAN~\citep{hu2019see}, to segment sub-moves. Considering these sports characteristics, we then construct a motion embedding function for each sub-motion to capture the structural similarities. 

\chenNew{To build the motion embedding space, we follow the successful parametric model, Skinned Multi-Person Linear model (SMPL) \citep{SMPL:2015}, which represents the pose parameters (rotation vectors) as $\boldsymbol{\theta}$. However, different from SMPL or other parametric pose/shape models \citep{romero2017embodied,SMPL-X:2019}, we propose the pose coefficient $\boldsymbol{\alpha}$ (see Fig.~\ref{fig:feature}) to leverage the rich semantic and temporally structural prior of sub-motions in the motion embedding space.} Specifically, for a sub-motion $m$, the motion embedding function is formulated as follows:
\label{motion embedding}
\begin{align}
  &\boldsymbol{\theta} = \mathcal{M}_m(\boldsymbol{\alpha}),\\
  &\boldsymbol{\theta} = \sum_{k=1}^K\alpha_k\mathbf{b}^m_k + \mathbf{a}^m = \boldsymbol{\alpha}^\intercal\mathbf{B}^m + \mathbf{a}^m,
  \label{motion embedding function}
\end{align}
where $\mathcal{M}_m(\boldsymbol{\alpha}): \mathbb{R}^K \mapsto \mathbb{R}^{3N}$, $N$ denotes the joint number of SMPL, and $K$ denotes the dimension of pose coefficients $\boldsymbol{\alpha}$. $\mathbf{a}^m$ is the mean of pose parameters and $\boldsymbol{\alpha}=[\alpha_1,\cdots,\alpha_K]^\intercal$ are the pose coefficients. Although the poses of the sports are challenging, the similarity of the poses in a sub-motion shows a desirable feature on lower-dimension. Thus, we adopt the Principal Component Analysis (PCA) to model the pose space of each sub-motion. We collect a MoCap dataset with more than 50 thousand frames to provide the set of pose parameters $\{\boldsymbol{\theta}^i\}$ and conduct PCA on $\{\boldsymbol{\theta}^i\}$ to generate a set of pose bases $\mathbf{B}^m=\{\mathbf{b}^m_k\}_{k=1}^K$, so that $\boldsymbol{\theta}$ under sub-motion $m$ can be represented as a linear combination of the bases. With our MoCap data, we calculate all pose bases $\{\mathbf{B}^m, \mathbf{a}^m\}$ for each sub-motion $m$ before our training parts.

We not only formulate the motion embedding function $\mathcal{M}_m(\boldsymbol{\alpha})$ on pose parameters $\boldsymbol{\theta}$ in Eq. \ref{motion embedding function}, but also formulate it on 2D or 3D joints as $\mathbf{J} = \mathcal{M}'_m(\boldsymbol{\alpha})$. \chenNew{We use Fig.~\ref{fig:feature} to visualize the motion embedding spaces on 2D joints, 3D joints, and pose parameters, namely $(\mathbf{J}_{2D}, \mathbf{J}_{3D}, \boldsymbol{\theta})$. It describes the variation on first three pose bases $\{\mathbf{b}^m_k\}_{k=1}^3$ in two sub-motions of diving and boxing, where \textit{mean} denotes $\mathbf{a}^m$ of these pose variables, for example, \textit{+0.5 std} on \textit{PC1} indicates $\boldsymbol{\alpha} = \{0.5, 0,...,0\}$ in Eq.~\ref{motion embedding function}.} The approach reduces the dimension of poses, benefiting for training, and regression. It can also robustly handle all sub-motions even for traditionally challenging poses because of the extracted structure of pose spaces. The resulting pose coefficients representation also encodes semantic meanings (see Fig. \ref{fig:Coeff_Vis}), like rotation angle, important for subsequent action parsing Sec.~\ref{action attributes}.

From the pose bases for all sub-motions, we construct the Motion Embedding Module that estimates the pose coefficients and 3D joints. This motion embedding representation can be suitable for many kinds of backbones. In our case, Motion Embedding Module consists of a ResNet-152 convolutional encoder followed by two fully connected layers to regress the pose coefficients $\boldsymbol{\alpha}$ used to reconstruct the joint positions:
\begin{align}
  \boldsymbol{\alpha}(\mathbf{x}) & = \mathcal{F}^m_{conv}(\mathbf{x};\mathbf{W}),                           \\
  \mathbf{J}(\mathbf{x})          & = \boldsymbol{\alpha}(\mathbf{x})^\intercal\mathbf{B}^m + \mathbf{a}^m,
\end{align}
where $\mathbf{x}$ denotes input image/frame and $\mathcal{F}^m_{conv}$ is the motion embedding network for the sub-motion $m$. of shape capturing. We utilize this network to estimate pose coefficients, shape parameters and camera parameters (see Eq. \ref{loss_data}) from images. Then, we recover 3D human body meshes from estimated pose and shape parameters of SMPL, the parametric human model.

Unlike prior approaches that target at general poses by implicitly encoding pose regularity into a complex network and hence cannot 
easily enforce semantic constraints, Motion Embedding Module manages to exploit the structural constraints in sport poses with action semantics. Even though we formulate the motion embedding function (Eq. \ref{motion embedding function}) with the pose parameters of SMPL, Motion Embedding Module can also be applied to other joint representations, like 2D/3D joint location/rotation, which is used in our experiment of 2D pose estimation (Fig. \ref{fig:PoseResults}). 

\subsection{Action Parsing}
\label{action attributes}
We then provide the estimated pose coefficients and 3D joint positions of all frames from Motion Embedding Module to the Action Parsing Module for analyzing the complete action. It includes inferring semantic meaningful labels and the action number(code) from the image sequence of sports, and later assessing the performance.

\chenNew{Specifically, we introduce \label{sas} Semantic Attributes (SAs) to represent the semantic meaningful label assumed by the sport guidelines/rules. The action number represents a valid combination of SAs, an overall description of a sports action. For example, as shown in Fig. \ref{fig:DiveIntro}, the five SAs for a diving action are the take-off type, twisting number, somersault number, arm-stand, and dive position, while the specific action number consists of are a set of these SAs.}

The brute-force approach would be to build a black-box network to map the pose sequence to the action number. Competitive sports have detailed defined elements and fixed semantic attribute types. So, we adopt a different approach that treats the action number as attributes of the action. Recall the action number encodes critical semantic meaning of the sport move in Fig.~\ref{fig:DiveIntro}. We call them semantic attributes (SAs) and aim to learn how each frame contributes to respective attributes. Our Action Parsing Module explicitly recovers SAs via a two-stage architecture: in the first, we use a multi-Stream ST-GCN for SA predictions, and in the second, a attributes mapping block infers the action number or score from the SAs.

\begin{figure}[t]
  \centering
  \includegraphics[width=0.8\linewidth]{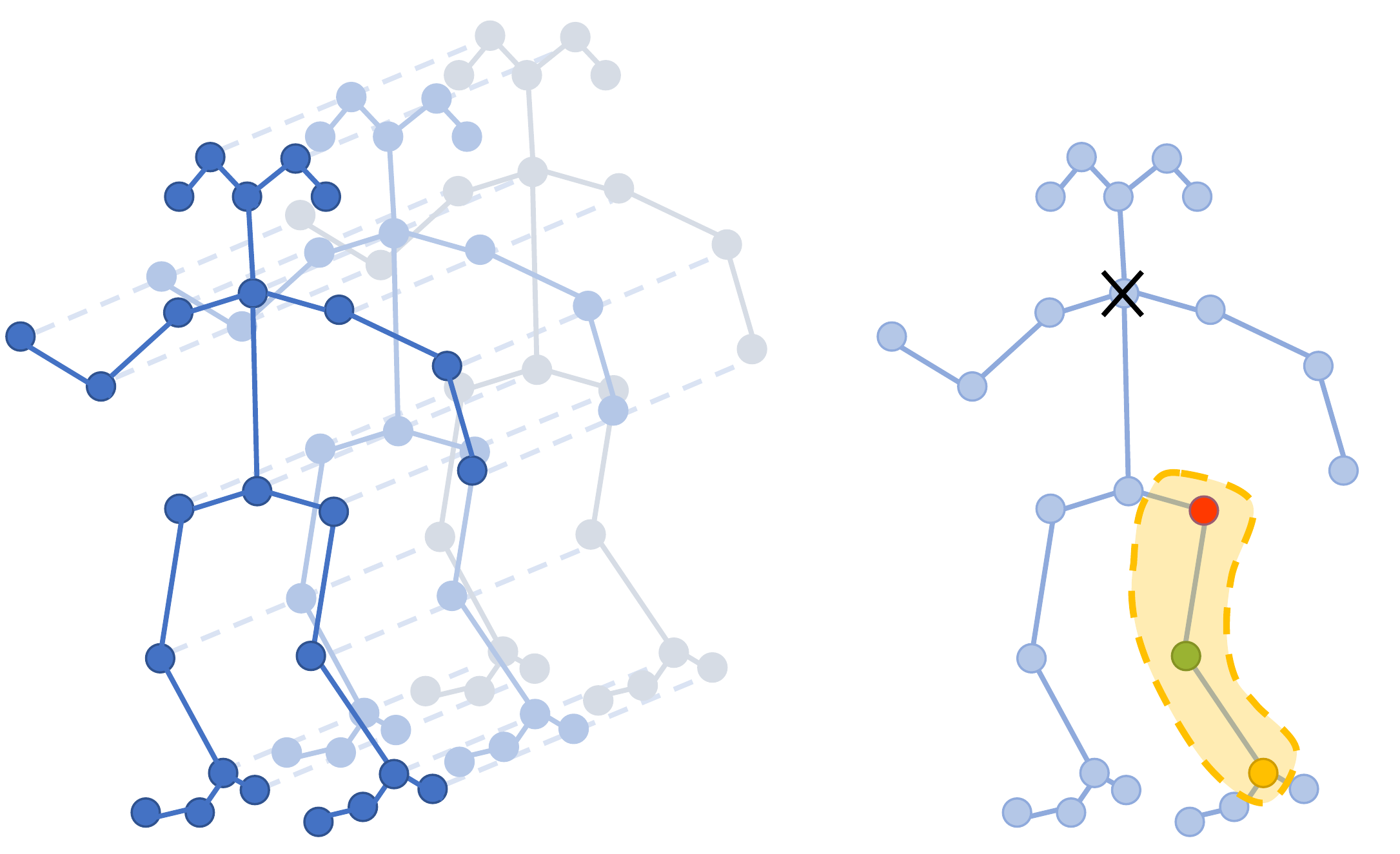}
  \caption{We show our spatial-temporal graph of the joints and bones. Right shows the spatial configuration partitioning strategy: we divide the node's one neighbor into three subsets: the root node (green dot), the centripetal subset (red dots), and the centrifugation subset (yellow dots), details in \citet{li2018spatio}.}
  \label{fig:GCN_graph} 
\end{figure}

\begin{figure}[tb]
  \centering
  \includegraphics[width = \linewidth]{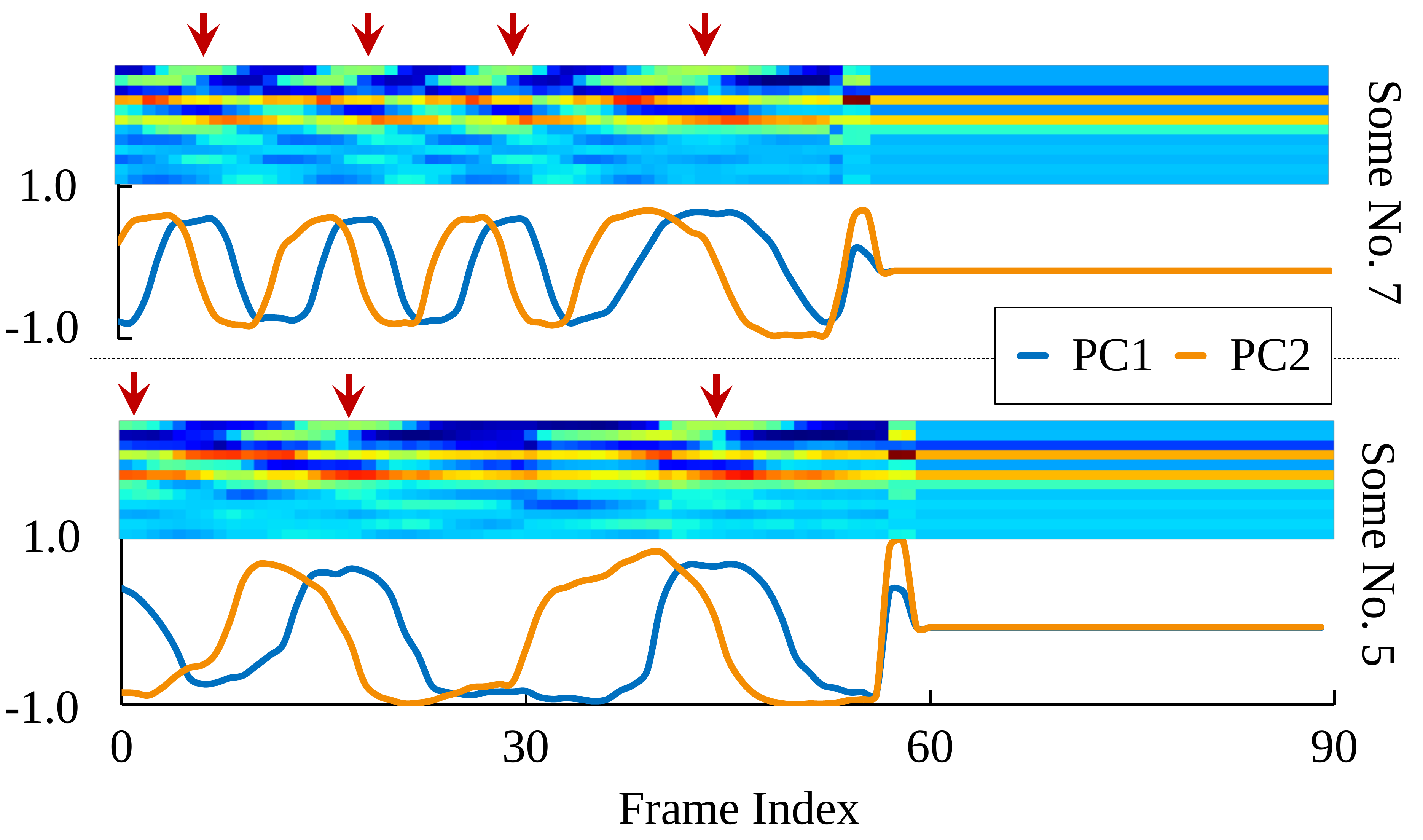}
  \caption{Extracted feature maps of pose coefficients from the P-stream encode the semantic meaning of sport. Each feature map is of 90$\times$25 (90 frames and 25-dimensional features)\pang{In each figure, the upper half is the principal components feature map of the entire sequence, and the lower half is the specific numerical curve of the first two principal components.}. The first two principal components already reveal the number of rounds of half-somersault (left: $4\times2-1=7$ rounds; right: $3\times2-1=5$ rounds). Red arrows correspond to the start position of a somersault (toe pointing to the ground), whereas the final half-somersault corresponds to entry to the water.}
  \label{fig:Coeff_Vis}
\end{figure}

\myparagraph{Spatial-Temporal Feature Extraction.} A number of previous approaches such as \citep{wen2019graph,si2018skeleton,zhang2019graph} exploit skeletons alone as inputs to the GCN. \pang{Our pose coefficients encode meaningful semantic sub-motion using proposed motion embedding analysis}. In addition to skeletons (bones and joints), pose coefficients obtained from the Motion Embedding Module provide useful information on action parsing, as shown in Tab.~\ref{tab:action}. We thus construct a multi-Stream convolutional module that takes joints (J-stream), bones (B-stream), and pose coefficients (P-stream). For J- and B-Stream, we adopt the 2s-AGCN structure that can adaptively learn graph edge connections. Details on graph construction and partitioning are shown in Fig.~\ref{fig:GCN_graph}.  Specifically, we adopt the human joints and bones setting in OpenPose~\citep{cao2018openpose}. In the J-stream, the joints are mapped to graph nodes, and the bones map to edges. In the B-Stream, the mappings are reversed. Note that we feed 90 consecutive frames of skeletons into a 10-layer ST-GCN to extract two feature vectors. For the P-stream, we represent pose coefficients as a 1D vector and use layers of 1D convolution with residual blocks to generate features of 256 dimensions. We demonstrate the effectiveness of the proposed P-Stream. In Fig.~\ref{fig:Coeff_Vis}, we visualize one of the feature maps of a specific sequence obtained from the P-Stream, at a resolution of 90$\times$25 (90 frames and 25 dimensions in feature). We also plot the first two dimensions vs. frame index. We observe that they can be readily used to infer the somersault number of competitive diving. We finally concatenate all feature vectors generated by the J-, B- and P-Stream as inputs to the following attributes mapping block.

\myparagraph{Semantic Attributes Mapping Block.}
The Semantic Attributes Mapping Block aims to learn the mapping between the extracted spatial-temporal features and the final action label, i.e., to tell which dive number or action scores the video corresponds to. Instead of directly learning the mapping via a black-box solution, we sought to use Semantic Attributes (SAs) explicitly. Specifically, our goal is to partition the whole action sequence in terms of the SAs, or more precisely, how individual frames contribute to specific SAs. We use two fully connected layers to predict their contributions where the categories of all SAs are represented as vectors. Finally, we stack the resulting SAs and feed the results to another two fully connected layers to infer the action number. Compared with black-box end-to-end approaches, our results show the use of intermediate SAs better supervise the training process, provide heuristic cues analogous to human labeling, and accelerate the training process. Moreover, decomposing the whole action into the SAs resembles human perception, which helps analyze sports videos' fine-grained actions.

\subsection{Multi-task Training}
\label{Training Strategy}
To train our end-to-end multi-task network, we adopt a deeply-supervised strategy to design five losses for the Motion Embedding Module and Action Parsing Module.

\myparagraph{Loss for Motion Embedding Module.} In our network, we use three types of representations, pose coefficients, pose parameters and 3D joints, to model poses. The pose coefficients define the motion embedding space of the pose, whereas the joint positions better describe the visibility between joints within an image. We therefore design the prior loss (the pose coefficients loss) $\mathcal{L}_{prior}$ as:
\begin{align}
  \mathcal{L}_{prior} = \left\|\mathbf{W}(\overline{\boldsymbol{\alpha}} -\hat{\boldsymbol{\alpha}})\right\|_2,
\end{align}
where $\overline{\boldsymbol{\alpha}}$ is the mean of pose coefficients in training set.
$\hat{\boldsymbol{\alpha}}$ is the predicted pose coefficients. Then, $\mathbf{W}$ is the weights, which calculates from the eigenvalues of a covariance matrix in the calculation of pose embedding bases. We use the smaller weight for the larger eigenvalue to enable more tolerance on principal components. We design the data loss (the 2D joint position loss) $\mathcal{L}_{data}$ as:
\begin{align}
    \label{loss_data}
    &\hat{\mathbf{J}}=\hat{s} \Pi(\mathcal{J}(\mathcal{M}_m(\hat{\boldsymbol{\alpha}}), \hat{\boldsymbol{\beta}}))+\hat{t}, \\
    &\mathcal{L}_{data} = \left\|\mathbf{V} (\mathbf{J}-\hat{\mathbf{J}})\right\|_2,
\end{align}
where $\mathbf{J}$ is the ground truth 2D joints and $\mathbf{V}$ indicates the visibility of the ground truth joint. Pose parameters $\hat{\boldsymbol{\theta}} = \mathcal{M}_m(\hat{\boldsymbol{\alpha}})$ is recovered from our embedded pose coefficients $\hat{\boldsymbol{\alpha}}$, using this motion embedding function of the sub-motion $m$. 3D joints $\mathcal{J}(\hat{\boldsymbol{\theta}}, \hat{\boldsymbol{\beta}})$ are obtained by linear regression from the final mesh vertices of SMPL. We then follow \citet{HMR} using a weak-perspective project system with only scale $s$, translation parameters $t, t \in \mathbb{R}^2$, and the orthographic projection function $\Pi(\cdot)$. We design the SMPL loss as:
\begin{align}
  \mathcal{L}_{smpl} = \left\|\mathbf{\boldsymbol{\theta}}-\hat{\boldsymbol{\theta}}\right\|_2 + \left\|\mathbf{\boldsymbol{\beta}}-\hat{\boldsymbol{\beta}}\right\|_2,
\end{align}
where $\mathbf{\boldsymbol{\theta}}$,$\mathbf{\boldsymbol{\beta}}$ are the supervision of pose/shape parameters, which are obtained through MoSh \citep{loper2014mosh} and provided mocap data.

We combine these three loss terms, the prior loss of our motion embedding, the data loss, and the pose/shape parameter loss of SMPL with the corresponding weights $\lambda_{data}$, $\lambda_{smpl}$ (10 and 2 in our case) as the final loss of the Motion Embedding Module:
\begin{align}
  \mathcal{L}_{mem} = \mathcal{L}_{prior} + \lambda_{data}\mathcal{L}_{data} + \lambda_{smpl}\mathcal{L}_{smpl}.
\end{align}

\myparagraph{Loss for Action Parsing Module.}
For the Action Parsing Module, we use the cross-entropy loss between the predicted and the ground truth attributes, which can be written as follows,
\begin{align}
  \label{cross entropy}
  \mathcal{L}_{attr}= \sum^{Ns}_{c=1} \sum_{i=1}^{N_c} y_{i}^{c} \log \left(x_{i}^{c}\right),
\end{align}
where $c$ indicates the attribute type, $N_s$ denotes the number of attributes,
and $N_c$ is the categories of each attribute $c$.For example, the number of somersault in diving has 10 categories, indicating the rotation angle for 0 to 3240 ($0*360$ to $9*360$) degrees.
Here $x_i^c$ denotes the prediction for the $i$-th label of attribute $c$ whereas $y_i^c$ is the ground-truth.

For the action labeling task, we also add the cross-entropy loss between the prediction and the ground truth action label as below. We note that such a task loss depends on the target application and can be easily adjusted according to the final task.
\begin{align}
  \mathcal{L}_{task}= \sum_{j=1}^{N_f} y_{j}\log \left(x_j\right),
\end{align}
where $N_f$ denotes the total number of all possible action labels (action number or score), and $x_j, y_j$ are the prediction and ground truth of the $j$-th final label.

The overall loss for the Action Parsing Module is a combination of the attribute loss and the task loss as following:
\begin{align}
  \mathcal{L}_{apm} = \mathcal{L}_{attr}  + \lambda_{A} \mathcal{L}_{task},
\end{align}
where $\lambda_{A}$ (2 in our case) is a weight to balance two loss terms.

\myparagraph{Training Strategy.} While our entire network can be trained in an end-to-end fashion, we exploit its modular architecture and develop a stage-wise strategy, which is more efficient in practice. Specifically, our training procedure is composed of three stages: 1) We train a Motion Embedding Module for each sub-motion independently and fix its parameters, 2) We then train the action attribute prediction and label classification modules in the Action Parsing Module jointly, and 3) Finally we fine-tune the entire network using the combined losses of Motion Embedding Module and Action Parsing Module (i.e., $\mathcal{L}_{
mem}+\mathcal{L}_{apm}$).
\label{dataset}

\begin{figure*}[t]
  \centering
  \includegraphics[width=\linewidth]{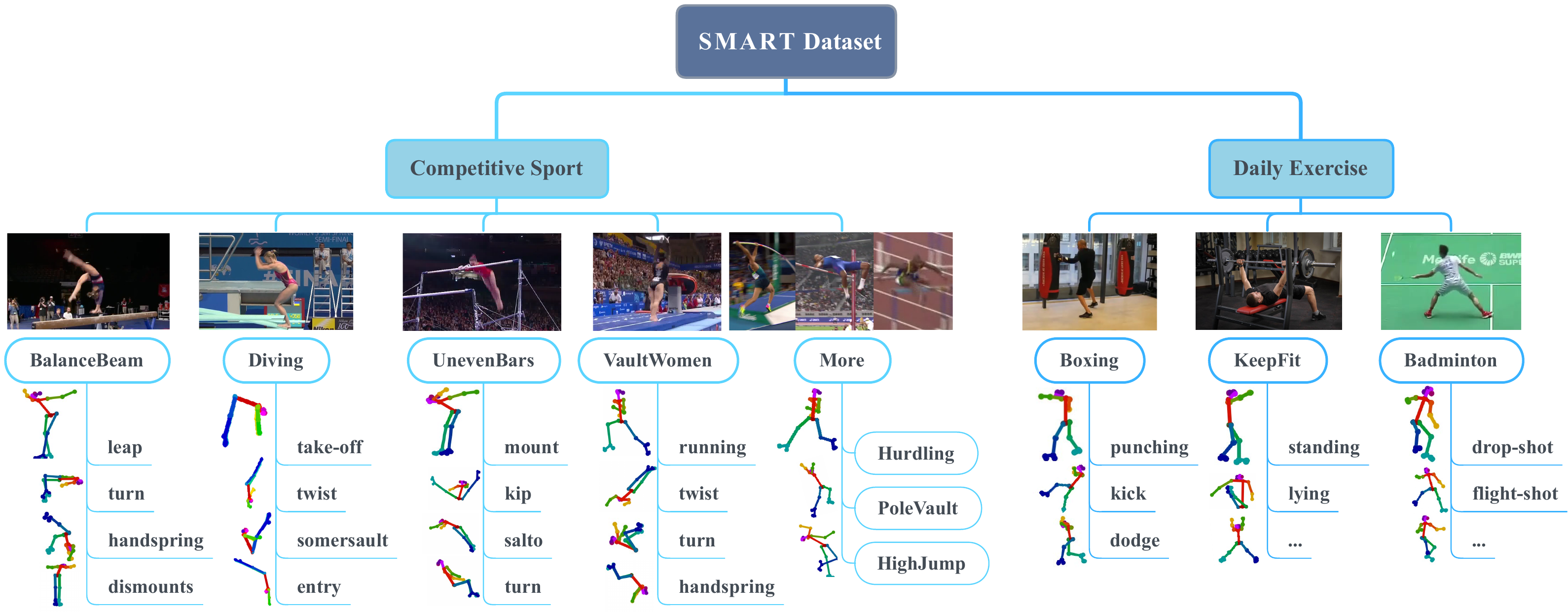}
  \caption{Our SMART dataset contains both pose and action annotation. The upper part shows specific categories of sport. The lower part depicts the sub-motions of each sport and its typical poses.}
  \label{fig:dataset_intro}
\end{figure*}
\section{Experimental Results}
\label{ExperimentalResults}
In this section, we evaluate our SportsCap in a variety of challenging scenarios.
We first report a new proposed dataset and the training details of our approach on the utilized datasets, followed by evaluating our main technical components, including both qualitative and quantitative comparisons with previous state-of-the-art methods on both motion capture and action parsing tasks.
Finally, limitations and discussions regarding our approach are provided.

\subsection{Dataset}
\begin{figure}[t]
  \centering
  \includegraphics[width=\linewidth]{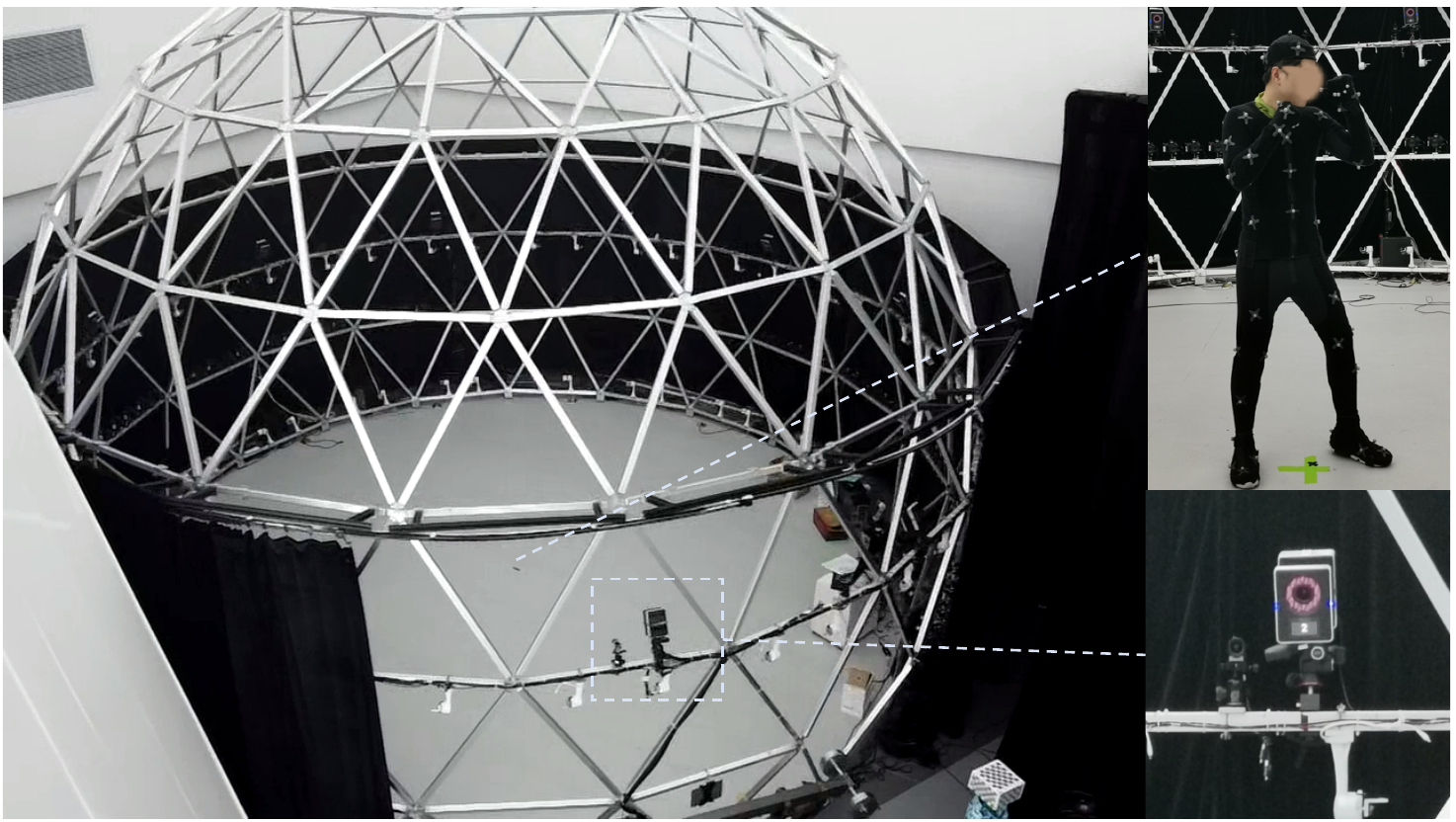}
  \caption{The motion capture system. The structure of dome is shown in the left and the motion capture camera is shown in the bottom right. Our system includes 12 multi-view motion capture cameras. The top right figure shows the motion cap suit with 63 marks. With professional performers, we utilize this system to capture 3D challenging motion of sports.}
  \label{fig:mocap_system}
\end{figure}

Many datasets exist for human action analysis, such as the MPII~\citep{andriluka14cvpr} and COCO~\citep{lin2014microsoft} for human pose in daily activities, the PennAction~\citep{Zhang_2013_ICCV} for coarse sport recognition, and the AQA~\citep{Parmar_2019_CVPR} and FineGym~\citep{shao2020finegym} dataset for fine-grained action recognition.
We propose a challenging sports dataset called Sports Motion and Recognition Tasks (SMART) dataset, which contains per-frame action labels, manually annotated pose and action assessment of various challenging sports video clips from professional referees.
We also collect the human pose data in sports activities using a marker-based motion capture system (see Fig.~\ref{fig:mocap_system}) to provide pose prior to our Motion Embedding Module.
To our best knowledge, our SMART dataset is the most complete dataset for human motion capture and action analysis for sport video (see Tab. ~\ref{tab:dataset}).

\begin{figure*}[ht]
  \centering
  \subfigure[Sub-motions of competitive diving.]{
    \centering
    \includegraphics[width=0.4\linewidth]{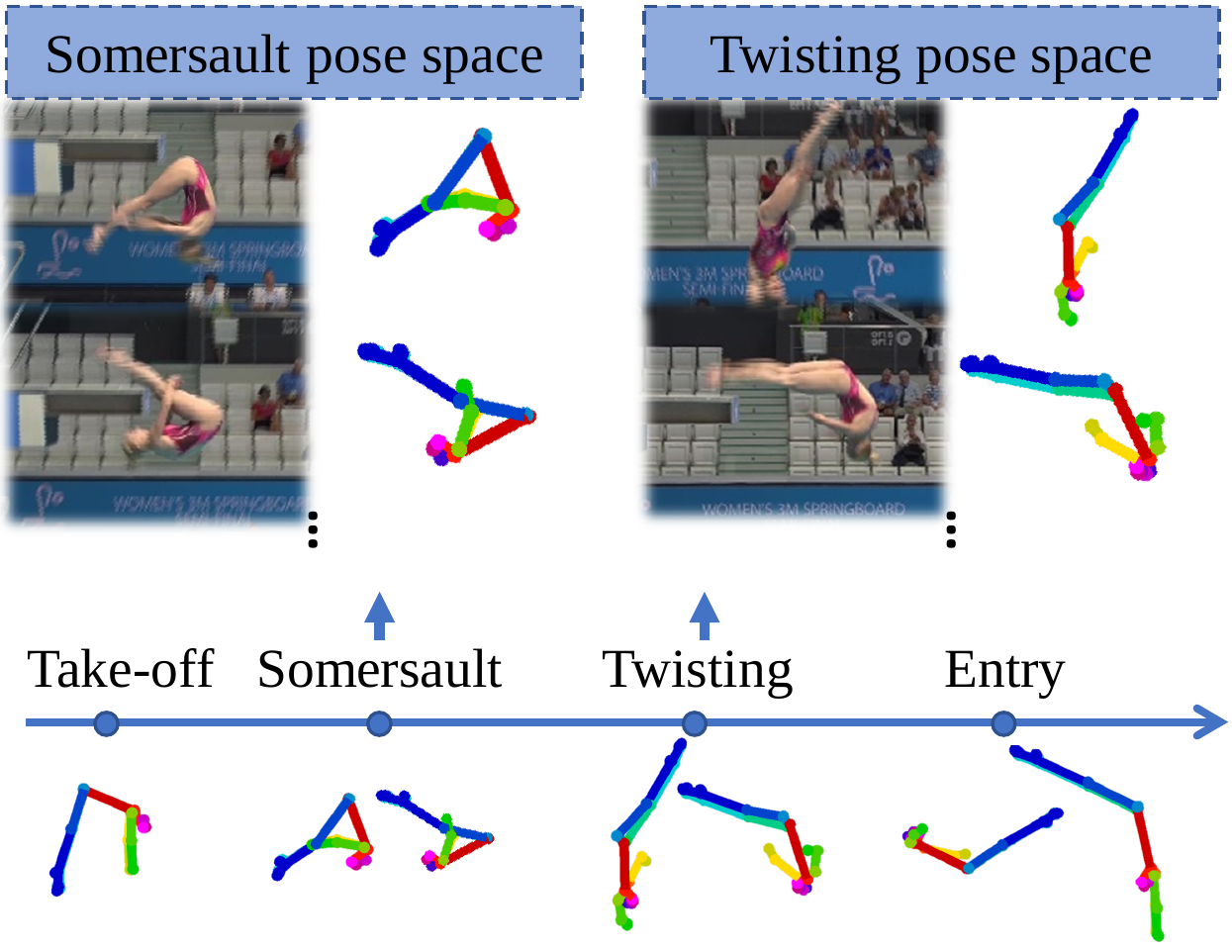}
    \label{fig:DivePoseSpace}}
  \subfigure[Semantic attributes of competitive diving.]{
    \centering
    \includegraphics[width=0.48\linewidth]{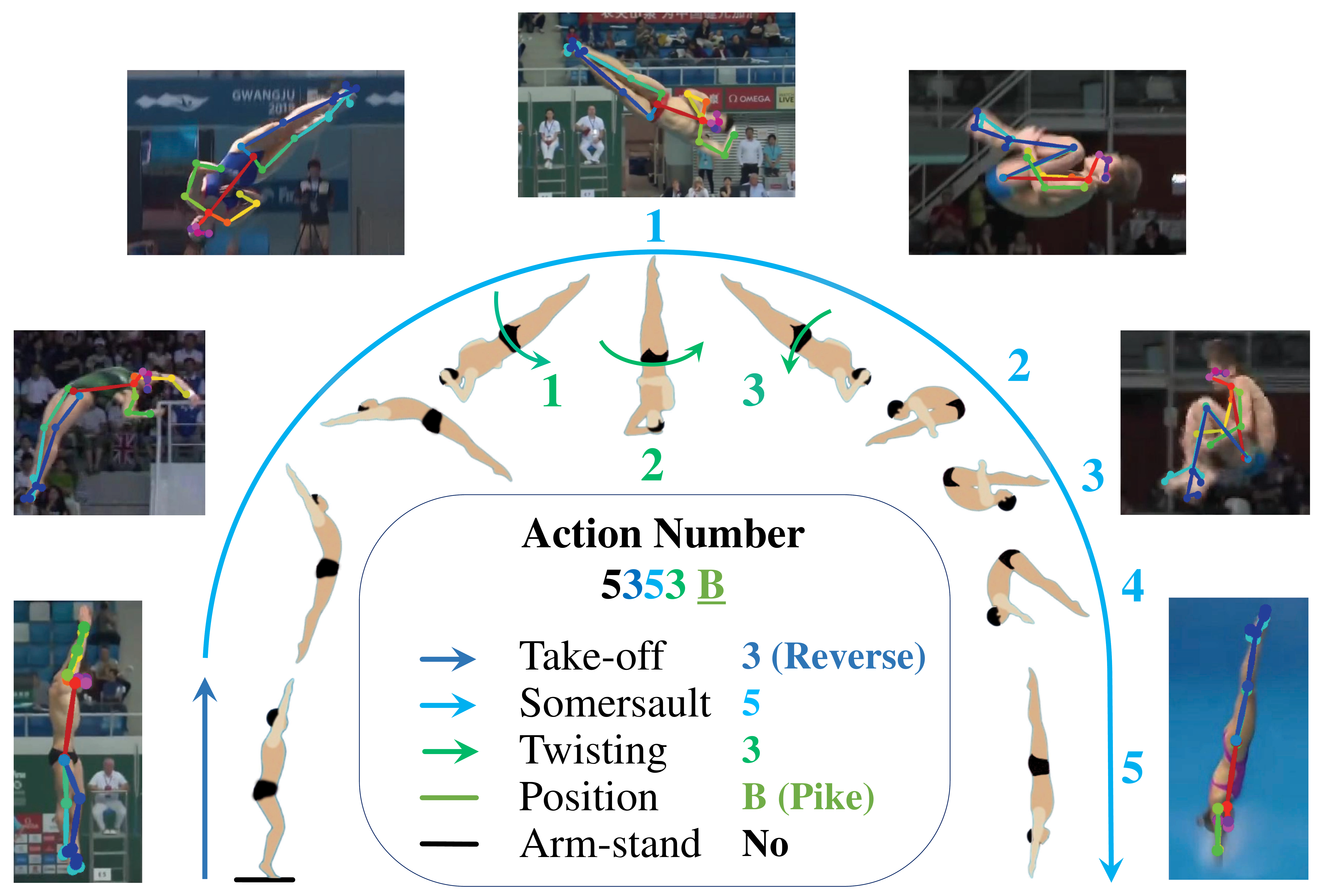}
    \label{fig:DiveIntro}}
  \caption{Two important types of label annotations in the SMART dataset, sub-motions (SMs) and semantic attributes (SAs). We introduce SAs and SMs of competitive diving as example. (a) SMs indicate the different pose spaces, usually different stages in a sport, like somersault and twisting in this case. (b) SAs indicate the specific number of a motion, like the rotation angle, take-off type and so on. }
\end{figure*}

Our SMART dataset consists of both competitive sports and daily exercise videos (see Fig. \ref{fig:dataset_intro}), including balance beam, competitive diving, uneven bars, vault-women, hurdling, pole vault, and high jump in competitive sports, and boxing, keep-fit, and badminton in the daily exercise category.
The SMART has 640 videos (110K frames) in total, with per-frame skeleton annotation and sub-motion labels, semantic attribute labels, and action assessment scores for competitive sports.
There are about 450,000 annotated skeletons, 25 joints like OpenPose\citep{cao2018openpose}, with corresponding bounding boxes.
In addition to joint locations, we also annotate the visibility of each joint as three types: visible, labeled but not visible, and not labeled, same as COCO \citep{lin2014microsoft}.
To fulfill our goal of 3D pose estimation and fine-grained action recognition, we collect two types of annotations, i.e. the sub-motions (SMs) and semantic attributes (SAs), as we described in Sec. \ref{sms}/\ref{sas} and Fig. \ref{fig:DivePoseSpace}/\ref{fig:DiveIntro}.
We also include the difficulty scores, the number of valid referees, the execution scores, and the final assessment scores for competitive sports in the SMART dataset.
\chenNew{The action labels (not include joints) of gymnastics sports in SMART dataset are from FineGym \citep{shao2020finegym}. All the other annotations, including joints and action labels, are manually generated with professional cross-validation between more than two individuals to guarantee the annotation accuracy. We will share our SMART dataset with the community.}

\begin{table}[t]
  \caption{Comparison between the SMART dataset and existing datasets, including MPII-Pose~\citep{andriluka14cvpr}, Penn Action~\citep{Zhang_2013_ICCV}, COCO~\citep{lin2014microsoft}, AQA~\citep{Parmar_2019_CVPR} and FineGym~\citep{shao2020finegym}, regarding size, per video action labels, pose annotation, action assessment, and pose type. We show the distribution and annotation details of each specific sport of SMART in the bottom of the table. The bold means the best performance or attribute. }
  \centering
  \label{tab:dataset}
  \begin{spacing}{1.4}
    \resizebox{\linewidth}{!}{
      \begin{tabular}{lccccccc}
        \hline
        Dataset       & \makecell[c]{Video                                                                                                                     \\Clips} & Images & \makecell[c]{Actions \\ per clip} & Pose & \makecell[c]{3D\\mocap} & Assess & Type \\
        \hline
        MPII          & -                  & 25K           & -              & \checkmark          & $\times$            & $\times$            & General        \\
        COCO          & -                  & 330K          & -              & \checkmark          & $\times$            & $\times$            & General        \\
        Penn          & 2K                 & 160K          & Single         & \checkmark          & $\times$            & $\times$            & Sport          \\
        AQA           & 1K                 & 150K          & Multi          & $\times$            & $\times$            & \checkmark          & Dive           \\
        FineGym       & 5K                 & 1.9M          & Multi          & $\times$            & \checkmark          & $\times$            & Gymnastics     \\
        \textbf{Ours} & \textbf{5K}        & \textbf{2.1M} & \textbf{Multi} & \textbf{\checkmark} & \textbf{\checkmark} & \textbf{\checkmark} & Sport \\
        \hline
        -             & 640                & 110K          & Multi          & \checkmark          & 15K                 & \checkmark          & Dive           \\
        -             & 2K                 & 434K          & Multi          & \checkmark          & 5K                  & \checkmark          & Vault-Women    \\
        -             & 1K                 & 490K          & Multi          & \checkmark          & 5K                  & \checkmark          & UnevenBars     \\
        -             & 1K                 & 704K          & Multi          & \checkmark          & 2K                  & \checkmark          & BalanceBeam    \\
        -             & -                  & 24K           & Single         & \checkmark          & 5K                  & $\times$            & Hurdling       \\
        -             & -                  & 155K          & Single         & \checkmark          & 2K                  & $\times$            & PoleVault      \\
        -             & -                  & 96K           & Single         & \checkmark          & 3K                  & $\times$            & HighJump       \\
        -             & -                  & 4K            & Single         & \checkmark          & 11K                 & $\times$            & Boxing         \\
        -             & -                  & 54K           & Single         & \checkmark          & 2K                  & $\times$            & Badminton      \\
        \hline
      \end{tabular}}
  \end{spacing}
\end{table}

\begin{figure*}[htb]
  \centering
  \includegraphics[width=\linewidth]{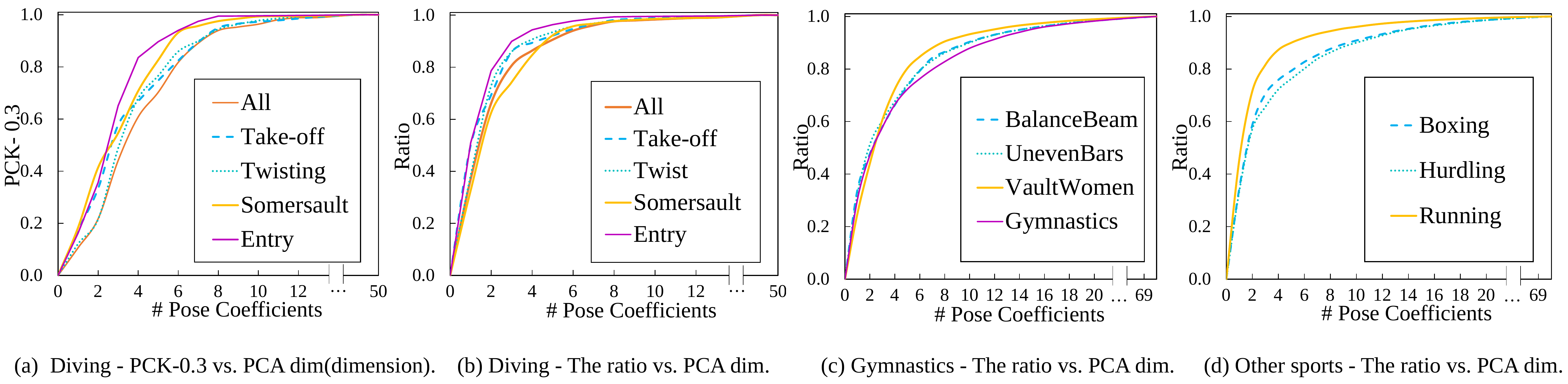}
  \caption{Cumulative relative variance of our sports dataset explained as a function of the number of pose coefficients. (a) and (b) are the sub-motions of diving with different metrics. ``All'' refers to use all training data rather than restricted to each semantic pose space. (c) and (d) are general motion embedding of gymnastics and daily sport. PCKs indicates the percentage of correct keypoints (see details in~\citep{andriluka14cvpr}), and the ratio follows \citep{SMPL:2015}.}
  \label{PCK_Chart}
\end{figure*}

In addition to the annotated video data, we also collect the 3D human pose data using a motion capture system to provide pose motion prior for our Motion Embedding Module.
Specifically, we adopt the Vicon system (a 12 views marker-based motion capture system) to capture the rich human pose sequences and hire 30 athletes and two professional fitness instructors as our performers.
The performers move according to the corresponding sport guidelines to make sure their body movements cover as many sub-motions of the sports as possible (except the motions that are impossible to execute in the capture environment).
For the generalization purpose, each performer repeats the movements two to five times, and the same motion is captured from more than two different subjects.
Since only the relative motion matters, we convert the skeleton results from Vicon to the SMPL pose parameters to avoid the variations imposed by the absolute lengths of bones.
In total, we collect more than 500,000 motion frames of 30 performers covering nine activities, about 1,000 frames for each performance.

With the annotated 2D poses and MoCap 3D pose data, we collect the Sports Motion Embedding Spaces according to our motion embedding function (Eq.~\ref{motion embedding function}) and use it as the pose priors for sports videos. Currently, Sports Motion Embedding Spaces provides the priors on 2D joints, 3D joints location, and pose parameters of SMPL for nine sports, as shown in Tab. \ref{tab:dataset}, and we are planning to add more sports in the near future. Because of the regularity of the human body motion in sports/exercises, the Sports Pose  Embedding  Spaces provides strong prior and regularization to ensure that the generated pose result lies in the corresponding action space. The Sports Pose Embedding  Spaces greatly improves the accuracy and robustness of the 2D/3D pose estimation, human body capturing, action recognition/parsing, and action assessment tasks as described in Sec.~\ref{Motion Embedding Module}.  The Sports Pose Embedding  Spaces data will be included in the SMART dataset.

\subsection{Training Details}
Our Motion Embedding Module relies on both the fine-grained action labels and pose information. Therefore, we first train and test the Motion Embedding Module on the SMART dataset as few other datasets provide both information. Then we fix the Motion Embedding Module and train the complete SportsCap on SMART, AQA~\citep{Parmar_2019_CVPR} and FineGym dataset~\citep{shao2020finegym} for 3D sports motion estimation and action understanding.

We resize image patches that contain the human body at a resolution of 256$\times$256 (using the ground truth bounding box in our SMART dataset and detect the bounding box in AQA \citep{Parmar_2019_CVPR} using \citet{liu2016ssd}). We re-sample the video to 90 frames each. For Motion Embedding Module training, we conduct data augmentation via random rotations ($-45^{\circ}$ to $+45^{\circ}$), random scaling (0.7 to 1.3), and flipping horizontally. For Action Parsing Module training, we also augment the skeleton and pose parameters data for J-, B- and P-Streams, respectively. For J-Stream and B-Stream, we scale the joint positions via interpolation to simulate the far and near camera views. For the P-Stream, we also scale the coefficients vector.

We use the Adam optimizer \citep{kingma2014adam}, train the Motion Embedding Module in the first 100 epochs and AMP in the following 50 epochs. We train the complete SportsCap in the last ten epochs. The learning rates of the 0th, 70th, and 150th epoch are $10^{-3}$, $10^{-4}$, and $10^{-5}$, respectively. We train our SportsCap on 4 NVidia 2080Ti GPUs, and the process takes 10 hours for the Motion Embedding Module, 5 hours for Action Parsing Module, and 2.5 hours for the whole SportsCap. Once trained, the network processes the $90\times256\times256$ video data at 0.5s for Motion Embedding Module, 0.05s for Action Parsing Module, and 1.0s for the data fetching.

For fair comparisons, we re-train HRNet \citep{sun2019deep} and SimpleBaseline \citep{xiao2018simple} using our SMART dataset. For SCADC \citep{Parmar_2019_CVPR}, C3D-LSTM \citep{parmar2017learning}, C3D-AVG \citep{Parmar_2019_CVPR}, and R2+1D \citep{tran2018closer}, we first pre-train the corresponding networks using the UCF101 dataset \citep{soomro2012dataset} and I3D \citep{sun2019deep} on the Kinetics dataset \citep{sun2019deep}, replace their output or the regression layers with our proposed Semantic Attributes Mapping Block module, and fine-tune Semantic Attributes Mapping Block with our SMART dataset.

\begin{figure*}[ht]
  \centering
  \includegraphics[width=\linewidth]{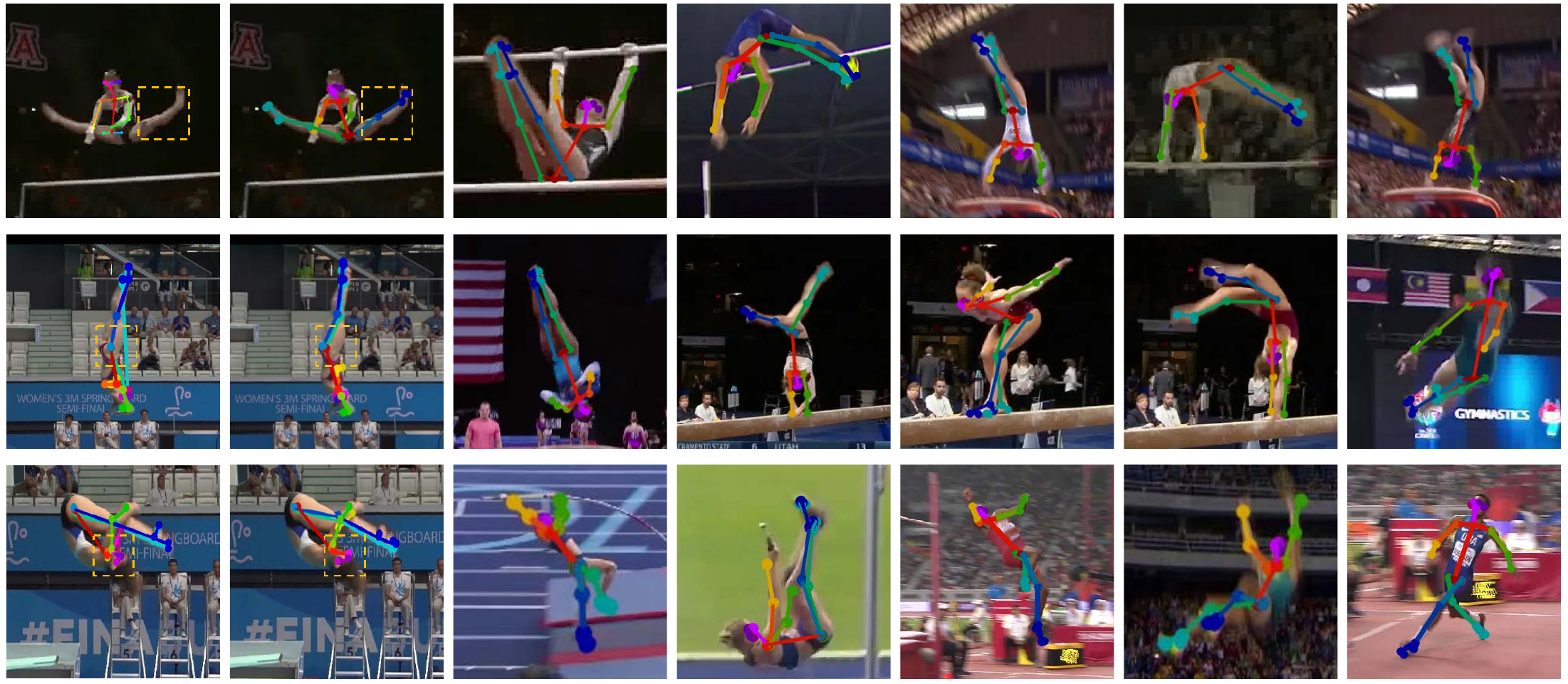}
  \caption{The gallery and comparison of our experiments on 2D pose estimation. With 2D pose embedding spaces, we show the comparison with fine-tuned HRNet \citep{sun2019deep} (The first column) on SMART dataset. Our results (start from the second column) are more robust and reliable under challenging poses and motion blur.}
  \label{fig:PoseResults}
\end{figure*}

\begin{figure*}
  \centering
  \includegraphics[width=0.82\linewidth]{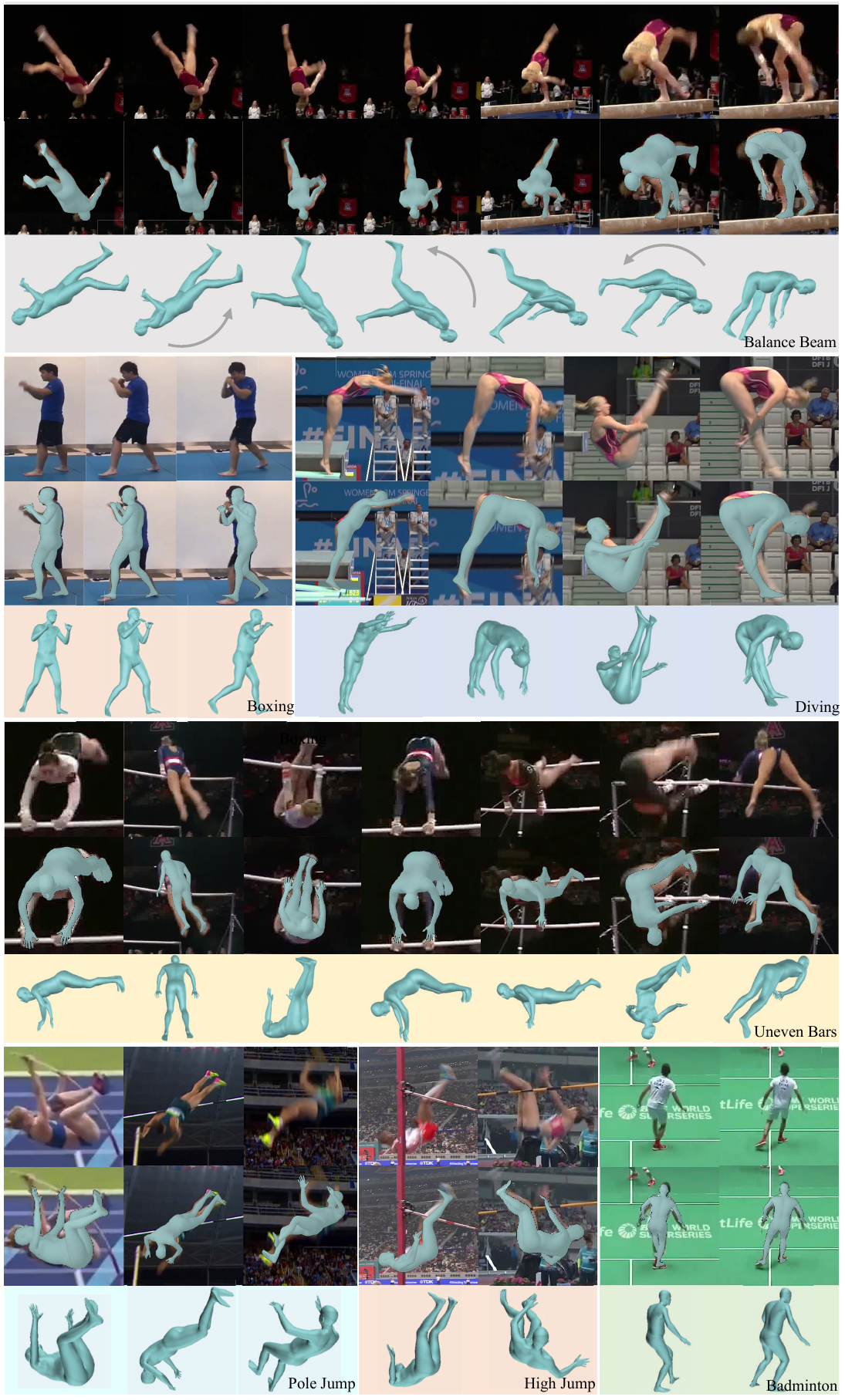}
  \caption{3D human shape recovery results on challenging sport videos. For each type of sport, the top row shows the input images, the middle row shows the recovered body mesh, and the bottom row shows the rendering result of the recovered body from an alternative view.}
  \label{fig:rs_gallery}
\end{figure*}

\begin{figure}[t]
  \centering
  \includegraphics[width=\linewidth]{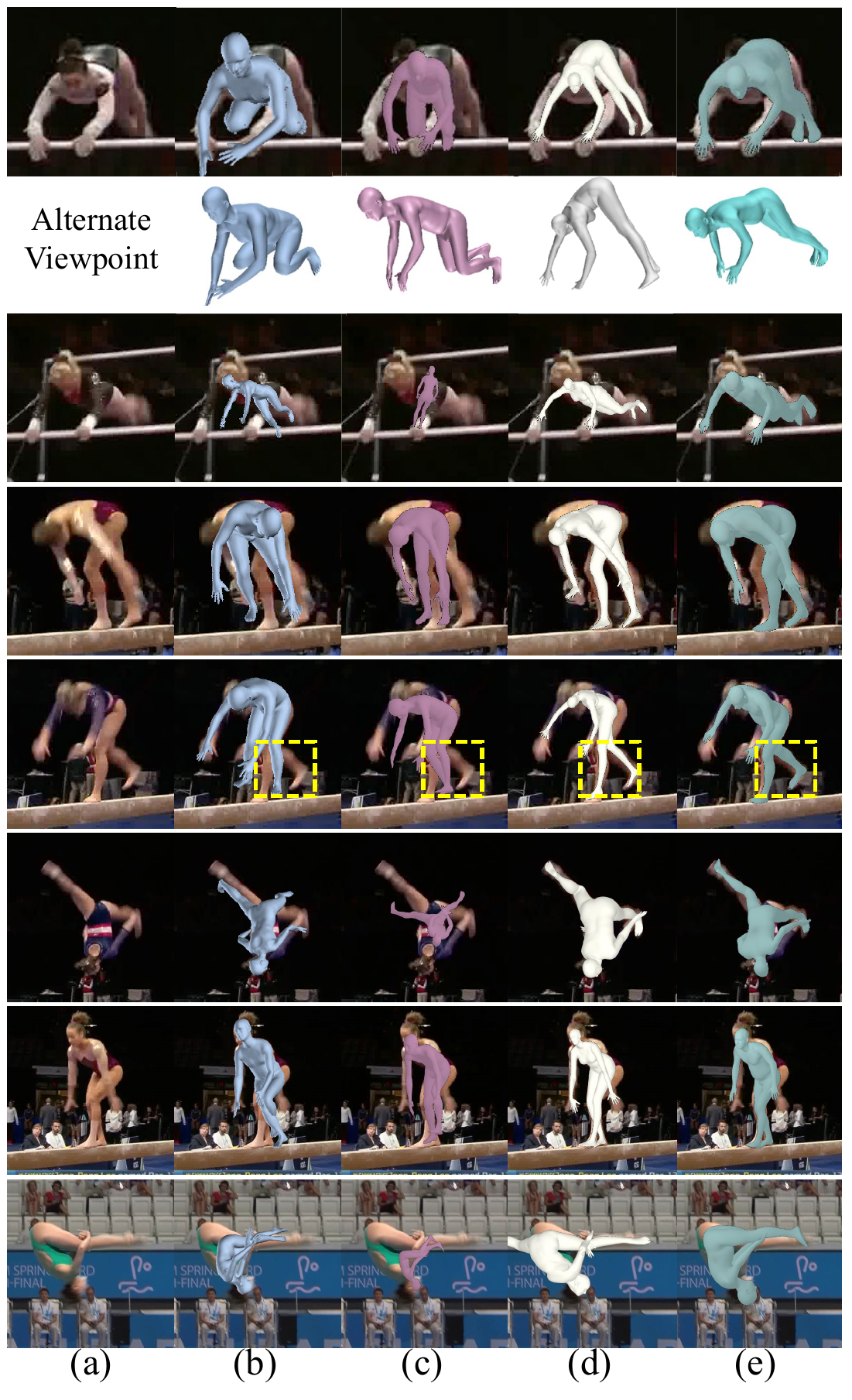}
  \caption{Qualitative comparison with the state-of-the-art human shape recovery methods. (a) The input images. (b), (c), (d) are the results of HMR \citep{HMR}, VIBE \citep{kocabas2020vibe}, and SMPLify-X \citep{SMPL-X:2019} respectively.  We fine-tune HMR and VIBE on our training set, and provide the 2D poses of our method for SMPLify-X. (e) Our results of SportsCap On SMART dataset.}
  \label{fig:comp_body}
\end{figure}

\subsection{Evaluation on 3D Motion Capture}
In this sub-section, we evaluate our SportsCap approach with three motion-relevant tasks, including the motion embedding module evaluation, comparison of motion capture, and the sub-motion classification.

\begin{table}[t]
  \centering
  \caption{Ablation study of our Motion Embedding Module: PCA w/o sub-motion labeling, training w/o multi-task learning (w/o action parsing module), 50/101/152 ResNet as Backbone, and $\mathcal{L}_{prior}$/$\mathcal{L}_{data}$ as loss for training. With no $\mathcal{L}_{prior}$, we follow \citet{HMR} use pose parameters as the variable of $\mathcal{L}_{data}$ directly on the SMART dataset. We use the percentage of correct keypoints (PCK-0.3, PCK-0.5) as our metrics.}
  \label{tab:pose}
  \resizebox{\linewidth}{!}{
    \begin{tabular}{l|c|c|c|cc}
      \hline
      TrainingLoss                               & Task   & PoseSpace & Backbone  & PCK-0.3       & PCK-0.5       \\
      \hline
      $\mathcal{L}_{data}$                       & Multi  & -         & ResNet50  & 70.9          & 86.5          \\
      $\mathcal{L}_{prior}$                      & Multi  & Semantic  & ResNet50  & 83.6          & 91.5          \\ \hline
      $\mathcal{L}_{prior}$+$\mathcal{L}_{data}$ & Multi  & General   & ResNet50  & 83.0          & 92.1          \\
      $\mathcal{L}_{prior}$+$\mathcal{L}_{data}$ & Multi  & Semantic  & ResNet50  & 84.8          & 92.4          \\
      $\mathcal{L}_{prior}$+$\mathcal{L}_{data}$ & Multi  & Semantic  & ResNet101 & 87.5          & 94.5          \\
      $\mathcal{L}_{prior}$+$\mathcal{L}_{data}$ & Single & Semantic  & ResNet152 & 88.1          & 94.8          \\
      $\mathcal{L}_{prior}$+$\mathcal{L}_{data}$ & Multi  & Semantic  & ResNet152 & \textbf{88.5} & \textbf{96.0} \\
      \hline
    \end{tabular}
  }
  \vspace{1pt}
\end{table}
\myparagraph{Motion Embedding Module evaluation.} \chenNew{There are two ablation results, Fig.~\ref{PCK_Chart} and Tab.~\ref{tab:pose}. The first experiment verifies that it is more effective to use motion embedding analysis for different pose spaces by comparing PCA analysis in different pose spaces. Another experiment validates our Motion Embedding Module's specific designs and the mutual gain between 3D motion capture and fine-grained motion understanding. It illustrates that mutual gain will make the 2D keypoint projection results more accurate.

In Fig.~\ref{PCK_Chart}, PCA analysis demonstrates that the poses of each sub-motion lay in a low-dimensional parametric space, which is similar to the low-dimensional shape space in SMPL \citep{SMPL:2015}. In Fig.~\ref{PCK_Chart}(a)/(b), we use two metrics, the relative cumulative variance ratio and PCK-0.3, to evaluate semantic/general pose spaces from the training and testing sets, respectively. Hence, under the same dimension of pose coefficients, each semantic pose space's pose spaces show better accuracy than the general pose space. This proves that motion embedding analysis on each sub-motion is necessary and effective. Thus, compared with conducting PCA on the complete action data, we further reduce the space dimension by cooperating with the different sub-motion labels.

In Tab.~\ref{tab:pose}, we evaluate our Motion Embedding Module quantitatively from the ablation study on the SMART dataset. We project the 3D joints with camera parameters and consider the predicted results with distance errors less than 0.3 and torso length errors less than 0.5 as the correct predictions (see details in~\citep{andriluka14cvpr}) and report the percentage of correct keypoints (\textbf{PCK-0.3} and \textbf{PCK-0.5}) as our metrics. We train Motion Embedding Module without the prior loss provided by Sports Pose Embedding Spaces (with $\mathcal{L}_{prior}$), which is an approximately 5\% drop in accuracy. We also train this module w/o Action Parsing Module (\textbf{Multi} or \textbf{Single}), which is above 1.2\% improvement on PCK-0.5 from this multi-task learning. To follow the PCA analysis on sub-motion poses in Fig.~\ref{PCK_Chart}, we also use \textbf{General} or \textbf{Semantic} pose spaces in different trainings. We further test \textbf{ResNet-50}, \textbf{-101}, \textbf{-152} as our backbones for the encoder and find that about every 50 more layers lead to an above 2\% increase for PCK-0.3. This experiment demonstrates the effectiveness of the Sports Motion Embedding Spaces and the improvement of the fine-grained pose embedding spaces.
}


\begin{table}[t]
  \centering
  \caption{The comparisons with different methods trained on the same SMART dataset: HRNet \citep{sun2019deep}, SimpleBaseline \citep{xiao2018simple}, HMR \citep{HMR}, and VIBE \citep{kocabas2020vibe} of our Motion Embedding Module. We project the recovered 3D joints from HMR and VIBE into image as their keypoint predictions. SM-1 to SM-4 are these sub-motions in Tab. \ref{tab_fineGrain}.} 
  \label{tab:comp_pose}
  \resizebox{\linewidth}{!}{
    \begin{tabular}{l|ccccc|c}
      \hline
      Method                                  & SM-1          & SM-2          & SM-3          & SM-4          & PCK-0.3       & PCK-0.5       \\
      \hline
      HRNet & 83.9 & 81.7          & 82.7          & 86.4          & 83.6          & 87.5          \\
      Simple& \textbf{86.3}          & 68.5          & 86.7          & 90.4          & 84.2          & 88.9          \\
      HMR& 81.3          & 68.9          & 71.9          & 73.1          & 73.8          & 84.1          \\
      VIBE& 72.0          & 41.9          & 78.7          & 59.4          & 44.1          & 62.4          \\
      Ours                                    & 83.6          & \textbf{84.6} & \textbf{91.5} & \textbf{94.0} & \textbf{88.5} & \textbf{96.0} \\
      \hline
    \end{tabular}
  }
\end{table}

\myparagraph{Comparison.}
\label{shape capturing experments}
\chenNew{
We first evaluate our approach with the qualitative results on pose estimation and 3D motion capture in Fig.\ref{fig:PoseResults} and Fig.\ref{fig:rs_gallery}, to show our generalization on various sports and environments. For qualitative comparison, in Fig.~\ref{fig:comp_body}, we compare our method with the state-of-the-art 3D shape recovery methods. We fine-tune \textbf{HMR} \citep{HMR} and \textbf{VIBE} \citep{kocabas2020vibe} with our dataset, and provide our results on 2D poses for SMPLify-X \citep{SMPL-X:2019}. Furthermore, we show both the recovered human shape rendered at the original and alternative viewpoints. As shown in Fig.~\ref{fig:comp_body}, for challenging sports movements, i.e., the handspring and somersault, both HMR and VIBE perform poorly while ours achieve much more accurate estimation. In contrast, our method has the ability to recover the limbs with higher fidelity than HMR and VIBE, as denoted by the colored box.
The SMPLify-X method provides reasonable 2D keypoint estimations but does not produce as good 3D shape results. Our approach outperforms all methods by generating both accurate 2D keypoints and well-matched 3D human shapes.

For quantitative comparison, we still use PCK-0.3 and PCK-0.5 as our metrics. Tab.~\ref{tab:comp_pose} shows the performance of the Motion Embedding Module compared with the state-of-the-art pose estimation work, including the HRNet~\citep{sun2019deep} and the SimpleBaseline~\citep{xiao2018simple}. We also compare with state-of-the-art 3D human shape recovery methods, including the HMR~\citep{HMR} and VIBE~\citep{kocabas2020vibe}. For HMR and VIBE, we project the joints of the recovered 3D human models into images and then use the projected joints as their predicted poses. For a fair comparison, we re-train the HRNet, SimpleBaseline, and et.al. on the SMART dataset. Note that we also provide the results of similar sub-motions (SM-1 to SM-4) from different sports to evaluate on various poses, especially these complex and challenging poses. Please refer to Tab. \ref{tab_fineGrain} for these specific sub-motions. 

As shown in Tab.~\ref{tab:comp_pose}, our approach outperforms the other baselines, especially on these challenging sport poses. HRNet~\citep{sun2019deep} and SimpleBaseline\citep{xiao2018simple} perform better on the performance of SM-1 (Take, Run, Mount in Tab. \ref{tab_fineGrain}), these more general poses, but can hardly handle other complex sub-motions. Similarly, VIBE and HMR perform better on SM-1, which is the preliminary motion of sports, while performing worse on others, which is the twist/ turning/ handspring motion. Twisting and other professional motions involve fast rotation and flipping of the body, while our proposed motion embedding from PCA has structure constraints on each sub-motion pose. With the real human poses as templates, our method achieves higher accuracy, leading by 4.3$\sim$4.9\% on PCK-0.3, and 7.1$\sim$8.5\% on PCK-0.5. It indicates our approach is more accurate in generating joint locations and the joints are more natural and stable.}



\begin{table}[t]
  \caption{Performance of the fine-grained sub-motion classification using the WS-DAN~\citep{hu2019see} on competitive diving video.}
  \centering
  \label{tab_fineGrain}
  \resizebox{\linewidth}{!}{
    \begin{tabular}{l|cccc|c}
      \hline
      Sports                       & SM-1     & SM-2  & SM-3 & SM-4       & Avg.                  \\
      \hline
      \multirow{2}{*}{Diving}      & Take     & Twist & Some & Entry      & \multirow{2}{*}{96.7} \\
                                   & 97.8     & 94.5  & 96.5 & 95.3       &                       \\
      \hline
      \multirow{2}{*}{VaultWomen}  & Run      & Twist & Turn & Handspring & \multirow{2}{*}{97.0} \\
                                   & 98.1     & 95.8  & 93.9 & 96.9       &                       \\
      \hline
      \multirow{2}{*}{BalanceBeam} & Dismount & Leap  & Turn & Handspring & \multirow{2}{*}{93.6} \\
                                   & -        & 94.6  & 91.3 & 95.0       &                       \\
      \hline
      \multirow{2}{*}{UnevenBar}   & Mount    & Kip   & Turn & Salto      & \multirow{2}{*}{96.7} \\
                                   & 93.2     & 96.7  & 97.5 & 94.6       &                       \\
      \hline
    \end{tabular}
  }
\end{table}

\myparagraph {Sub-motion classification.}
We also provide more evaluations of our sub-motion classifier on various sports. Notes that our Motion Embedding Module aims to embed the pose motion within a certain sub-motion of sports actions into parametric space. It relies on both the sub-motion labels and per-frame annotated pose for evaluation, we thus evaluate the classifier \citep{hu2019see} on predicting the sub-motion label. We observe this technique can achieve high accuracy, and the predicted sub-motion label helps the Motion Embedding Module for pose and shape recovery. \chenNew{Tab.~\ref{tab_fineGrain} shows our sub-motion classification produces an average accuracy around 96\% on various sports.}

\subsection{Evaluation on Fine-grained Action Understanding}

\chenNew{We evaluate the Action Parsing Module of the SportsCap and compare it with other state-of-the-arts on the SMART, AQA \citep{Parmar_2019_CVPR}, and FineGym \citep{shao2020finegym} datasets. These tasks include the fine-grained action parsing and the action assessment. 

\begin{table}[h]
  \begin{center}
    \centering
    \caption{Action parsing evaluation using state-of-the-art approaches vs. our method on the FineGym dataset \citep{shao2020finegym}.}
    \label{tab:FineGym}
    \resizebox{\linewidth}{!}{
      \begin{tabular}{c|l|c|c|c}
        \hline
        Dataset & Method      & VT            & UB            & partial Gym288 \\

        \hline
        \multirow{4}{*}{FineGym}
                & Random      & 16.7          & 6.7           & 0.3            \\
                & ST-GCN      & 19.5          & 13.7          & 11.0           \\
                & TRN-2stream & 31.4          & 83.0          & 42.9           \\
                & Ours        & \textbf{34.2} & \textbf{85.7} & \textbf{46.9}  \\

        \hline
      \end{tabular}
    }
  \end{center}
\end{table}

\begin{table*}[t]
  \begin{center}
    \centering
    \caption{Action parsing evaluation using state-of-the-art approaches vs. our Action Parsing Module method with joint/ joint+bone/ joint+bone+pose(coefficients) streams, on the SMART dataset and the AQA dataset \citep{Parmar_2019_CVPR}. \textit{SAMB} represents our approach with Semantic Attributes Mapping Block. }
    \label{tab:action}
    \begin{spacing}{1.1}  
    \resizebox{\textwidth}{!}{
      \begin{tabular}{c|l|ccccc|c}
        \hline
        Dataset                & Method                 & TakeOff       & ArmStand.     & Twist No.     & Some No.      & Position      & Diving No.    \\
        \hline
        \multirow{7}{*}{SMART} & C3D-LSTM               & 43.1          & 85.3          & 66.1          & 46.8          & 56.0          & 27.3          \\ 
                               & R2+1D                  & 34.9          & 84.4          & 64.2          & 44.9          & 55.6          & 26.1          \\ 
                               & I3D                    & 61.5          & 92.7          & 70.6          & 69.7          & 70.6          & 58.6          \\ 
                               & Ours (J)               & 85.3          & 97.5          & 78.0          & 82.9          & 75.6          & 65.0          \\
                               & Ours (J+B)             & 84.1          & 98.6          & 76.8          & \textbf{90.2} & 84.7          & 67.1          \\
                               & Ours (J+B+P) Black-Box & -             & -             & -             & -             & -             & 78.0          \\
                               & Ours (J+B+P)  SAMB     & \textbf{96.4} & \textbf{99.8} & \textbf{89.5} & 86.5          & \textbf{92.6} & \textbf{82.2} \\
        \hline
        \multirow{4}{*}{AQA}
                               & Nibali                 & 74.8          & 98.3          & 78.7          & 77.3          & 79.9          & -             \\ 
                               & MSCADC                 & 78.4          & 97.5          & 84.7          & 76.2          & 82.7          & -             \\ 
                               & C3D-AVG                & 96.3          & 99.7          & 97.5          & \textbf{96.9} & 93.2          & -             \\ %
                               & Ours                   & \textbf{97.5} & \textbf{99.8} & \textbf{97.9} & 96.3          & \textbf{94.0} & -             \\

        \hline
      \end{tabular}
    }
    \end{spacing}
  \end{center}
\end{table*}

\myparagraph{Fine-grained action parsing} \chenNew{In Tab. \ref{tab:FineGym}, we compare the SportsCap with the tested approaches in FineGym \cite{shao2020finegym}. Following their metrics and annotations, we test on the same fine-grained action labels with the mean accuracy. For annotations, e.g., \textit{Salto backward stretched with 2 twist} is decomposed to Salto backward, stretched, and 2 twist (three specific SAs), please refer to FineGym \citep{shao2020finegym} for the more details of ``annotating element labels''. We also use the motion embedding analysis under our 3D motion capture data. and please refer to FineGym \cite{shao2020finegym} for these compared approaches, Random in \citet{shao2020finegym}, ST-GCN \citep{yan2018spatial}, and TRN-2stream \citep{wang2018temporal}. It can be seen from the experimental results that our method performs 3-4\% improvement on the FineGym dataset than ST-GCN that also uses pose information. This is mainly because our 3D motion capture data and motion embedding analysis can better parse this type of sports motion under specific pose spaces.

In Tab. \ref{tab:action}, we first show the ablation study on the multi-stream structure and Semantic Attributes Mapping Block (SAMB) of Action Parsing Module. We evaluate all results with the Top-1 accuracy. Specifically, our multi-stream structure enables faster convergence. The 2s-AGCN structure (\textbf{J}- and \textbf{B}-Stream only) takes 70 epochs to converge, whereas the multi-stream structure converges after only 50 epochs. With \textbf{SAMB}, we further accelerate convergence to 10 epochs. For the effect of each stream, the result (Tab.~\ref{tab:action} row 4-6) shows the use of \textbf{P}-Stream significantly improves the accuracy vs. baseline (Tab.~\ref{tab:action} row 4, 5). Although the accuracy on the somersault attribute drops (Tab.~\ref{tab:action} row 5), this is expected as the network easily focuses only on one attribute without structure like SAMB. We further keep the multi-stream backbone but replace SAMB with \textbf{Black-Box} without using attribute loss $\mathcal{L}_{attr}$. The one with Black-Box converges after over 30 epochs with 78.0\% accuracy. In contrast, the network with SAMB converges after ten epochs with 82.2\% accuracy. }

We also compare the overall action parsing performance of diving with methods including \textbf{MSCADC} \citep{Parmar_2019_CVPR}, \textbf{C3D-LSTM}  \citep{parmar2017learning}, \textbf{C3D-AVG} \citep{Parmar_2019_CVPR}, \textbf{I3D} \citep{carreira2017quo} and \textbf{R2+1D} \citep{tran2018closer} in Tab. \ref{tab:action}. We still use the Top-1 accuracy as the metric. For the processing on the SMART dataset, we use the same strategy for all methods. Specifically, we sample 90 frames for a video clip. For multiple predictions, we also use multi-task blocks like ours. For AQA and FineGym dataset, we also sample each clip to 90 frames. Also, like them, we regard the fine-grained action recognition as a regression problem, to decompose all SAs to several to dozens of labels. Thus, we can compare with these approaches by regressing these labels. For the SMART dataset, SportsCap achieves the highest accuracy with 82.2\% Top-1 accuracy. It proves that using the pose coefficient from motion embedding benefits action parsing with P stream. For AQA dataset, we also compare SportsCap with Nibaili \citep{nibali2017extraction}. SMART achieves slightly better performance as C3D-AVG and outperforms MSCADC and Nabaili. It shows that our motion embedding method is effective not only on our dataset but also on other datasets with the same pose space.}

\begin{table}[t]
  \centering
  \caption{Action assessment comparisons on SMART and AQA dataset \citep{Parmar_2019_CVPR}, and compared with C3D-LSTM \citep{parmar2017learning} and R2+1D \citep{tran2018closer}.}
  \label{table:Assessment}
    \begin{spacing}{1.2}  
  \resizebox{\linewidth}{!}{
  \begin{tabular}{c|l|c|c}
    \hline
    Metric                    & Method   & SMART Dataset & AQA Dataset  \\
    \hline
    \multirow{3}{*}{Sp. Cor.} & C3D-LSTM & 53.7          & 84.9          \\
                              & R2+1D    & 55.6          & \textbf{89.6} \\
                              & Ours     & \textbf{61.7} & 86.2          \\
    \hline
  \end{tabular}
  }
  \end{spacing}
\end{table}
\myparagraph{Action Assessment.}
We further evaluate our approach for overall detailed action assessment using the diving motion, which relies on the dive number for final motion scoring.
We use spearman's rank correlation (\textbf{Sp. Cor.}) \citep{pirsiavash2014assessing} as the metric, and this action score can be regarded as a specific semantic attribute.
Specifically, instead of treating it as a regression problem, we discretize the score range 0-100 to 49 labels evenly. We use the cross-entropy loss Eq.~\ref{cross entropy} and conduct the same training strategy with the SA learning. We train our Motion Embedding Module module on our dataset for the poses, while training and testing Action Parsing Module for execution score and final score on our dataset and AQA dataset, respectively. In Tab.~\ref{table:Assessment}, we show the result of the testing results, 61.7\% on SMART and 86.2\% on AQA.


\subsection{Discussion}
As the first novel trial to explore the problem of joint 3D human motion capture and fine-grained motion understanding from monocular challenging non-daily video input, the proposed SportsCap still owns limitations. We list these discussions as follows.

\begin{figure}[t]
  \centering
  \includegraphics[width=\linewidth]{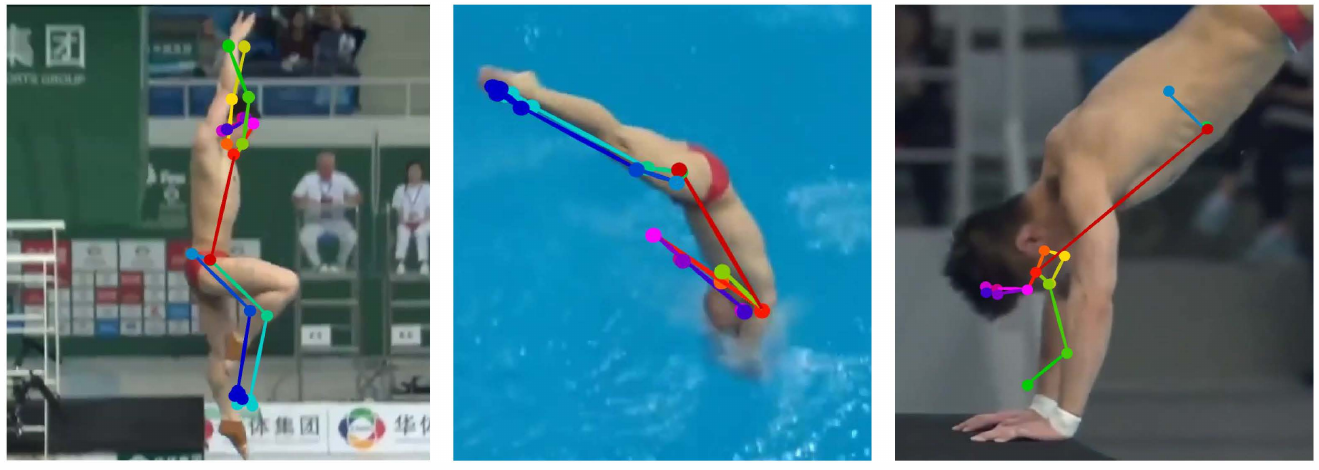}
  \caption{\chenNew{Failure cases. From left to right: (a) The input image does not fall into any of the pre-defined pose categories. (b) Some body parts are severely occluded and invisible. (c) The input image is incomplete due to clipping.}}
  \label{fig:failureCase}
\end{figure}

\myparagraph{Failure cases.} \chenNew{By tailoring our network for a specific subset of moves, our approach may generate erroneous estimations on degraded images. Fig.~\ref{fig:failureCase} provides several representative failure cases, where the input image does not fall into any of the pre-defined pose categories, some body parts are severely occluded and invisible (e.g., head entry into the water), or the input image is incomplete due to clipping. We plan to improve the motion embedding function with visibility and similarity parameters to handle invisible parts and unusual poses.}

\myparagraph{Limitations.} Our approach, in essence, exploits the semantic action analysis, for human pose estimation. This is different from HRNet \citep{sun2018integral} or SimpleBaseline \citep{xiao2018simple} that separately predict individual joints. Consequently, our network, once trained, only tackles specific sports rather than general movements as in prior art. In addition, our approach may generate erroneous estimations for large body parts outside the viewport. Moreover, our method can tolerate common deviations from the standard movement as we purposely add such cases into the dataset. However, when an athlete makes rare mistakes, it is difficult for our method to detect and analyze the situation accurately. \chenNew{Lastly, like many other works, SportsCap only estimates the motion of a single person for each inference. Although sub-motion is not a strict definition of motion semantics, it is not well suitable for team sports. Hence, the motion labels and spaces for multi-player sports might be necessary, to handle semantic interaction and multiple occlusions.}

\myparagraph{Future work}. Our current setup assumes a single video stream as input. In sports, it is common practice to show two or more video streams. In the future, we plan to combine multiple streams for a 3D pose/shape task. In addition, general human activities can always split into small sub-motions. Hence we plan to extend our work to general-purpose pose estimation through human action decomposition, including expanding our SMART dataset as a more general dataset. Besides, it's also an interesting direction to combine the NLP techniques to provide more natural and detailed illustrations and understanding for action assessment.

\section{Conclusions and Discussion}
We have presented the first approach for monocular markers-less 3D motion capture and understanding for professional non-daily motions and a new dataset consisting of various challenging sports video clips with rich manually annotated 2D/3D poses and
fine-grain action labels.
The key insight of our approach is to utilize the semantic and temporally structured sub-motion prior in the motion embedding space and formulate the joint motion capture and understanding task in a data-driven multi-task manner. 
%
Our motion embedding module achieves robust 3D motion details reconstruction from implicit motion embedding parameters,
while our novel multi-stream ST-GCN, as well as the semantic attribute mapping block, enable accurate fine-grained semantic action attributes prediction for various understanding applications like action assessment or motion scoring. 
Our experimental results demonstrate the effectiveness of SportsCap for both compelling 3D motion capture and fine-grained semantic action attribute reconstruction in various challenging sports scenarios, which compares favorably to the state-of-the-arts. 
We believe that it is a significant step to enable robust 3D motion capture and fine-grained understanding, with many potential applications in VR/AR, gaming, action recognition, and performance evaluation for gymnastics, sports, and dancing.

\begin{acknowledgements}
This work was supported by NSFC programs (\seqsplit{61976138}, \seqsplit{61977047}), the National Key Research and Development Program (\seqsplit{2018YFB2100500}), STCSM (\seqsplit{2015F0203-000-06}) and SHMEC (\seqsplit{2019-01-07-00-01-E00003}).
\end{acknowledgements}

{\small
\bibliographystyle{spbasic}
\bibliography{egbib}
}






\end{document}